%% file: main.tex
\documentclass{article}

\PassOptionsToPackage{table}{xcolor}

\usepackage{iclr2027_conference,times}
\usepackage{amsmath,amssymb}
\usepackage{amsthm}
\usepackage{booktabs}
\usepackage{graphicx}
\usepackage{float}
\usepackage{capt-of}
\usepackage{needspace}
\usepackage{enumitem}
\usepackage{microtype}
\usepackage{xcolor}
\usepackage{url}
\usepackage[
    colorlinks=true,
    linkcolor=red,
    anchorcolor=red,
    citecolor=brown,
    urlcolor=blue
]{hyperref}

\definecolor{oursrow}{HTML}{F4EBDD}

\newtheorem{proposition}{Proposition}

\newcommand{\method}{EviGDA}
\newcommand{\mstd}[2]{#1\,{\scriptsize$\pm$\,#2}}
\newcommand{\venue}[1]{{\scriptsize\textcolor{gray}{(#1)}}}

\title{Graph Domain Adaptation Does Not End with Representation Learning}

\iclrfinalcopy
\input{authors}

\begin{document}
\maketitle
\lhead{Preprint}

\input{arxiv/sections/00_abstract}
\input{arxiv/sections/01_introduction}
\input{arxiv/sections/02_related_work}
\input{arxiv/sections/03_method}
\input{arxiv/sections/04_experiments}
\input{arxiv/sections/05_results}
\input{arxiv/sections/06_discussion}
\input{arxiv/sections/07_statements}

\bibliography{arxiv/references}
\bibliographystyle{iclr2027_conference}

\clearpage
\appendix
\input{arxiv/sections/appendix}

\end{document}

%% file: arxiv/sections/00_abstract.tex
\begin{abstract}
Graph domain adaptation (GDA) transfers knowledge from a labeled source graph to an unlabeled target graph under shifts in both node attributes and graph structure.
Existing methods primarily adapt graph representations through propagation redesign, distribution alignment, or source-to-target transition modeling, but still rely on a single graph-propagating path for target prediction.
This leaves open whether an adapted graph representation exhausts the predictive evidence available in the target domain, since the graph-aware expert and graph-free local expert may exhibit different failure modes under topological shifts.
To address this limitation, we propose \textbf{EviGDA}, an Evidence-Augmented Graph Domain Adaptation framework that complements graph representation adaptation with a graph-free local expert.
The graph-aware expert performs message passing and entropy-aware marginal alignment, while the graph-free local expert learns solely from source node features and labels without graph propagation or target alignment.
The two experts are optimized independently and combined only at inference through a task-level constant probability mixture, preserving complementary evidence without joint training, learned routing, or target pseudo-labels.
Extensive experiments on ten datasets and 16 transfer tasks show that EviGDA
outperforms state-of-the-art baselines.

\end{abstract}

%% file: arxiv/sections/01_introduction.tex
\section{Introduction}
\label{sec:introduction}

Graph-structured data often exhibit distribution shifts across domains,
including changes in node attributes, class proportions, and neighborhood
relations. Models trained on a labeled source graph can therefore generalize
poorly to an unlabeled target graph. Graph domain adaptation (GDA) addresses
this problem by transferring knowledge across graphs with different attribute
and structural distributions
\citep{wu2020udagcn,wu2023grade,liu2023struRW,liu2024pairalign}. A predominant
line of work approaches GDA through representation adaptation. Source and
target embeddings are aligned by adversarial objectives or explicit
discrepancy measures \citep{wu2020udagcn,dai2023adagcn,wu2023grade,gretton2012kernel}; subsequent
methods account for conditional structure and label shifts
\citep{liu2023struRW,liu2024pairalign}, propagation and target smoothness
\citep{liu2024a2gnn,chen2025tdss}, or attribute, spectral, and homophily
discrepancies \citep{you2023specreg,fang2025attrgda,fang2025hgda,yang2025dgsda}.
Recent approaches learn adaptive alignment criteria or source-to-target
evolution \citep{chen2026adalign,chen2026diffgda}.

Despite these advances, most GDA methods ultimately infer target labels from a
\textbf{single graph-propagating prediction path}. Their improvements act on the
representation delivered to this path, leaving the final prediction restricted
to the information encoded after neighborhood aggregation. This restriction
may be consequential under structural shift: aggregation couples node
attributes to target-domain connectivity, whereas some attribute--label
relations may remain useful without that connectivity
\citep{liu2023struRW,liu2024pairalign,fang2025hgda,tai2026dft}. Representation
adaptation and predictive sufficiency are therefore different questions. Even
an improved graph representation need not exhaust the evidence available for
target prediction, while the role of residual evidence outside this
representation remains underexplored in GDA.

To address this limitation, we propose \textbf{\method}, an
Evidence-Augmented Graph Domain Adaptation framework that extends GDA beyond
representation adaptation. A graph-aware expert performs
domain adaptation using node attributes and connectivity, while a graph-free
local expert learns from labeled source attributes alone. The
two experts have disjoint parameters, objectives, and optimizers, and their
predictions are combined only after training through a task-level probability
mixture. The design keeps their information access distinct and permits a
direct test of whether local-expert predictions complement the adapted graph
expert. Within the graph-aware expert, we further introduce
entropy-aware exact-sampling alignment. It assigns nonzero sampling probability
to every target node while favoring lower-entropy predictions, thereby changing
target participation without changing the underlying multi-kernel discrepancy
estimator \citep{gretton2012kernel}. This change of empirical measure separates
node importance from discrepancy design and concentrates repeated alignment
estimates on more reliable target evidence.

To characterize the value of the local expert, we provide two complementary
results. First, the gap in minimum achievable log-risk between conditioning on
the adapted graph state and additionally conditioning on node attributes equals
their residual conditional information \citep{goldfeld2020information}.
Second, for a trained expert pair, an exact Brier-risk decomposition shows
that, when the graph-free local expert is weaker, some nonzero convex
mixture improves upon the graph-aware expert if and only if their expected
squared probabilistic disagreement exceeds their Brier-risk gap. The first
result identifies the information available beyond the adapted state; the
second determines whether the trained experts realize enough of that
difference to improve the mixture. Together, they formalize our central insight:
graph domain adaptation does not end with representation learning.

We evaluate \method{} in the standard 16-transfer setting across four
benchmark families used by recent GDA studies
\citep{liu2024a2gnn,chen2026adalign}. \method{} establishes a new state of
the art by outperforming existing baselines. Prediction-level
correctness decompositions reveal complementary errors between the two experts.
Under controlled topology corruption, degradation of the graph-aware expert is
accompanied by an increase in local-exclusive correctness.
Capacity- and ensemble-matched controls distinguish these gains from extra
parameters and same-expert averaging.

Our contributions are threefold:

\begin{itemize}[leftmargin=*]
\item We propose \method, an independent dual-expert framework
that complements graph representation adaptation with a graph-free local
expert to exploit evidence beyond the adapted representation.

\item We theoretically characterize when graph-free local
evidence improves an adapted graph expert and introduce entropy-aware
exact-sampling alignment to strengthen graph adaptation.

\item Extensive experiments on ten datasets and 16 transfer tasks show
that \method{} outperforms state-of-the-art baselines and consistently
benefits from complementary graph-free local evidence.
\end{itemize}

%% file: arxiv/sections/02_related_work.tex
\section{Related Work}

\paragraph{Representation-centric graph domain adaptation.}
Unsupervised domain adaptation commonly reduces the discrepancy between labeled source and unlabeled target representations through adversarial learning, moment matching, or kernel mean embedding
\citep{bendavid2010theory,ganin2016dann,gretton2012kernel,long2015dan,sun2016deepcoral,tzeng2017adda}.
Graph domain adaptation further needs to account for relational dependence and structural shifts.
Early cross-network approaches such as DANE and ACDNE combine transferable node encoders with adversarial distribution alignment
\citep{zhang2019dane,shen2020acdne}.
UDA-GCN and AdaGCN incorporate adversarial adaptation into graph encoders,
while GRADE characterizes non-IID graph transfer through representation-level
discrepancies \citep{wu2020udagcn,dai2023adagcn,wu2023grade}.
Subsequent methods make representation transfer sensitive to changes in graph structure:
StruRW reweights source neighborhoods under conditional structure shift,
Pair-Align jointly addresses conditional structure and label shifts,
A2GNN adapts propagation depth across source and target graphs,
TDSS improves target smoothness through sampled neighborhoods, and SpecReg
derives transfer-oriented spectral regularizers
\citep{liu2023struRW,liu2024pairalign,liu2024a2gnn,chen2025tdss,you2023specreg}.
Together, these methods progressively strengthen how transferable graph
representations are learned. The prediction endpoint itself has received less
attention: target decisions are generally derived from the resulting
graph-propagating representation. Our work examines complementary predictive
evidence retained by a separate graph-free local expert after graph
representation adaptation.

\paragraph{Decoupling graph shifts and adaptive transfer.}
Recent GDA methods further distinguish different sources or dynamics of domain shift.
Generative GDA disentangles semantic, domain, and nuisance factors
\citep{cai2024dgda};
GraphAlign instead modifies and compresses the source graph according to
alignment and rescaling principles \citep{huang2024graphalign};
SA-GDA performs category-aware spectral augmentation and combines local and
global graph views \citep{pang2023sagda};
GAA models attribute-driven transfer through interacting attribute and structural channels
\citep{fang2025attrgda};
HGDA explicitly aligns homophily with mixed graph filters
\citep{fang2025hgda};
JDA-GCN augments adversarial alignment with structural consistency
\citep{yang2024jdagcn};
and DGSDA separates attribute and topology adaptation through learnable spectral filters
\citep{yang2025dgsda}.
ADAlign learns an adaptive characteristic-function discrepancy, whereas DiffGDA models continuous structure--semantic evolution from source to target
\citep{chen2026adalign,chen2026diffgda}.
Most closely to our concern with neighborhood dependence, DFT decorrelates
features inside graph layers to reduce the conditional shift induced by local
dependencies \citep{tai2026dft}.
These methods ask which graph factors should be transferred and how they should
be aligned. We address a complementary question: whether the adapted graph path
should remain the sole prediction endpoint. Accordingly, our graph-free local expert
does not modify the graph encoder or construct an auxiliary graph; it preserves
a separate source-supervised view that never accesses adjacency. Within the
graph path, ADAlign adapts the discrepancy in spectral-frequency space, whereas
our EAM keeps multi-kernel MMD fixed and adapts the empirical target measure
through prediction-dependent exact sampling. It changes node importance in
alignment rather than introducing another representation view or discrepancy.

\paragraph{Multiple prediction paths and expert complementarity.}
Mixture-of-experts models learn dense or sparse input-dependent routing among specialized predictors, while deep ensembles aggregate independently trained models to improve predictive robustness
\citep{jacobs1991moe,shazeer2017moe,lakshminarayanan2017deepensembles}.
Under heterophily, H2GCN separates ego and neighbor embeddings to preserve
their distinct information \citep{zhu2020h2gcn}.
In graph learning, Mowst combines an MLP and a GNN through confidence-based node-wise cooperation, and GraphBridge augments a transferred GNN with a trainable side network
\citep{zeng2024mowst,ju2025graphbridge}.
\method{} instead studies post-adaptation evidence under unsupervised GDA.
It preserves two independently optimized experts with different information
access: one uses attributes and connectivity, while the other never performs
graph propagation. Unlike Mowst, their task-level mixture is not produced by a
node-wise router; unlike GraphBridge, the graph-free local expert is not trained
through a transfer, side-tuning, or fusion objective. The experts interact
only after training. Parameter-matched graph models and independently
initialized graph--graph ensembles then distinguish heterogeneous predictive
complementarity from capacity and generic ensembling.

%% file: arxiv/sections/03_method.tex
\section{Method}
\label{sec:method}

\begin{figure}[t]
    \centering
    \includegraphics[
        width=\textwidth,
        trim=28bp 92bp 6bp 64bp,
        clip
    ]{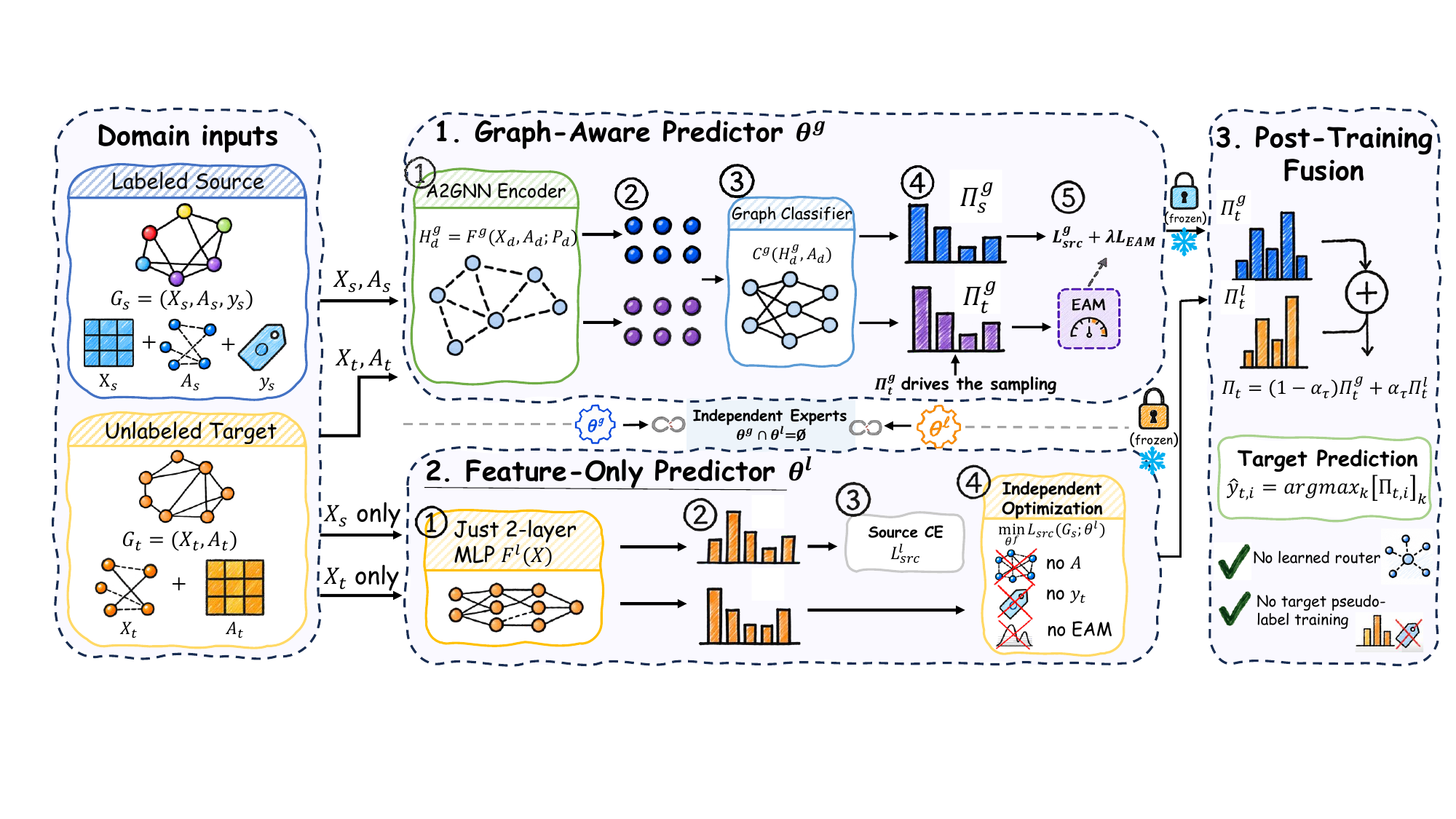}
    \caption{
        \textbf{Overview of \method.}
        The graph-aware expert is optimized by source classification and
        entropy-aware exact-sampling alignment. The graph-free local expert
        receives source supervision alone. Their parameters remain independent,
        and their target probabilities are combined after training using one
        coefficient shared by all nodes in the transfer task.
    }
    \label{fig:method}
\end{figure}

Let $\mathcal G_s=(\mathbf X_s,\mathbf A_s,\mathbf y_s)$ and
$\mathcal G_t=(\mathbf X_t,\mathbf A_t)$ denote the labeled source and
unlabeled target graphs, where $\mathbf X_d\in\mathbb R^{N_d\times D}$ and
$\mathbf A_d\in\mathbb R_{\geq0}^{N_d\times N_d}$ are the node attributes and
processed adjacency matrix of domain $d\in\{s,t\}$. The two domains share $K$
classes, and only the source labels
$\mathbf y_s\in\{1,\ldots,K\}^{N_s}$ are available during training.
Figure~\ref{fig:method} summarizes the graph-aware expert, independently trained
graph-free local expert, and post-training probability fusion in \method{}.

\suppressfloats[t]

\subsection{Graph-Aware and Graph-Free Local Experts}
\label{sec:method-overview}

We first obtain graph-aware node predictions using A2GNN
\citep{liu2024a2gnn}. Its encoder applies $P_d$ propagation steps in domain
$d$, allowing different source and target depths. Propagation uses symmetric
degree normalization with missing self-loops added. One feature transformation
and ReLU produce the embeddings, followed by one graph classification layer;
$P_d=0$ therefore removes encoder propagation but retains the graph classifier:
\begin{equation}
\mathbf H_d^g=F_g(\mathbf X_d,\mathbf A_d;P_d),
\qquad
\boldsymbol\Pi_d^g=C_g(\mathbf H_d^g,\mathbf A_d),
\qquad
\boldsymbol\Pi_d^\ell=F_\ell(\mathbf X_d).
\label{eq:graph-expert}
\end{equation}
Here $\mathbf H_d^g\in\mathbb R^{N_d\times h_g}$ contains the graph-aware node
embeddings, $\mathbf h_{d,i}^g$ denotes its $i$th row, and
$\boldsymbol\Pi_d^g,\boldsymbol\Pi_d^\ell\in\mathbb R^{N_d\times K}$ contain
class probabilities. The experts share source labels but deliberately differ
in information access: $F_g:(\mathbf X_d,\mathbf A_d)\mapsto\mathbf H_d^g$,
whereas $F_\ell:\mathbf X_d\mapsto\boldsymbol\Pi_d^\ell$ and therefore
$\partial\boldsymbol\Pi_d^\ell/\partial\mathbf A_d=\mathbf0$. Message passing composes
attributes with neighborhood evidence, which is useful when connectivity
transfers but can obscure stable attribute cues under structural shift. The
graph-free local expert is implemented as a feature-only two-layer MLP that
never reads $\mathbf A_d$.

The two experts also have disjoint parameters,
$\boldsymbol\theta_g\cap\boldsymbol\theta_\ell=\varnothing$, with
$\nabla_{\boldsymbol\theta_\ell}\boldsymbol\Pi_d^g=\mathbf0$ and
$\nabla_{\boldsymbol\theta_g}\boldsymbol\Pi_d^\ell=\mathbf0$. We denote their
source cross-entropies by $\mathcal L_{\mathrm{src}}^g$ and
$\mathcal L_{\mathrm{src}}^\ell$; the expert-specific objectives are given after
the alignment mechanism is defined. Their predictions interact only after
training, in probability space. A second head on $\mathbf H_d^g$ would inherit
the same adapted state and primarily test capacity or generic ensembling.
In contrast, $F_\ell$ preserves a separate attribute-based path;
Section~\ref{sec:capacity-controls} tests this distinction with controls for
parameter capacity and the choice of prediction path.

\Needspace{8\baselineskip}
\subsection{Why Representation Learning Need Not Be Sufficient}
\label{sec:method-insight}

Let $(X,Y,Z^g)\sim P_t$ describe a random target node, where $X$ and $Y$ are
its attributes and label and $Z^g$ is the state available to the graph
classifier after message passing. Target sufficiency requires
$P_t(Y\mid Z^g,X)=P_t(Y\mid Z^g)$ almost surely. For any input $S$, define
$\mathcal R_{\log}^{\star}(S)=\inf_q\mathbb E_t[-\log q(Y\mid S)]$.
For predictions, let
$\mathcal R(\boldsymbol\pi)=\mathbb E_t
\lVert\mathbf e_Y-\boldsymbol\pi\rVert_2^2$ be the Brier risk
\citep{gneiting2007proper}, for one-hot labels $\mathbf e_Y$. Set
$\mathcal R_b=\mathcal R(\boldsymbol\pi^b)$ for $b\in\{g,\ell\}$,
$\Delta=\mathcal R_\ell-\mathcal R_g$, and
$D=\mathbb E_t\lVert\boldsymbol\pi^g-\boldsymbol\pi^\ell\rVert_2^2$.

\begin{proposition}[Residual information and weak-expert gain]
\label{prop:residual-fusion}
The representation-level risk gap and the realized mixture risk satisfy
\begin{align}
\mathcal R_{\log}^{\star}(Z^g)-\mathcal R_{\log}^{\star}(Z^g,X)
&=I_t(Y;X\mid Z^g)\geq0,
\label{eq:residual-information}\\
\mathcal R(\boldsymbol\pi^\alpha)-\mathcal R_g
&=\alpha(\Delta-D)+\alpha^2D,
\quad \boldsymbol\pi^\alpha=(1-\alpha)\boldsymbol\pi^g+
\alpha\boldsymbol\pi^\ell.
\label{eq:fusion-risk-main}
\end{align}
The first equality is zero if and only if $Y\perp X\mid Z^g$. If
$\Delta\geq0$, some $\alpha\in(0,1]$ strictly improves the graph-aware expert if
and only if $D>\Delta$; then
\begin{equation}
\alpha^\star=\frac{D-\Delta}{2D},
\qquad
\mathcal R_g-\mathcal R(\boldsymbol\pi^{\alpha^\star})
=\frac{(D-\Delta)^2}{4D}>0.
\label{eq:optimal-fusion-gain}
\end{equation}
\end{proposition}

The first identity follows because Bayes log-risk equals conditional entropy
\citep{goldfeld2020information}; the second specializes the ensemble ambiguity
decomposition \citep{krogh1994ensembles} to two experts. Conditional information
identifies evidence beyond $Z^g$, while $D>\Delta$ tests whether the local expert
can improve a fixed mixture. If $D\leq\Delta$, $\alpha=0$ is optimal.
We evaluate Macro-F1 empirically and provide the complete proper-risk proofs
in Appendix~\ref{app:complementarity-theory}.

\subsection{Entropy-Aware Exact-Sampling Alignment}
\label{sec:eam}

Conventional marginal alignment samples target nodes uniformly and therefore
treats them as equally informative. Under domain shift, uncertain graph
predictions may reflect unstable neighborhood evidence and affect adaptation as
often as confident ones. EAM retains the kernel discrepancy but reallocates
sampling mass toward lower-entropy target nodes before measuring the
cross-domain discrepancy. For target node $i$, let
$\boldsymbol\pi_{t,i}^g$ be the $i$th row of $\boldsymbol\Pi_t^g$. Its
normalized predictive entropy and sampling probability are defined as follows.
The stop-gradient operator $\operatorname{sg}[\cdot]$ preserves forward values
but blocks backpropagation through the sampling probabilities.
\begin{equation}
e_i=-\frac{1}{\log K}\sum_{k=1}^{K}
\pi_{t,ik}^g\log\pi_{t,ik}^g,
\qquad
q_i=\operatorname{sg}\!\left[
\frac{\varepsilon+(1-\varepsilon)(1-e_i)}
{\sum_{j=1}^{N_t}[\varepsilon+(1-\varepsilon)(1-e_j)]}
\right].
\label{eq:entropy-sampling}
\end{equation}
We use $0\log0=0$ and a floor $\varepsilon\in(0,1]$ to give every target node
nonzero sampling probability. For $\varepsilon<1$, lower-entropy predictions
receive more mass; $\varepsilon=1$ recovers uniform sampling. The two laws
define entropy-tilted and uniform empirical target measures:
\begin{equation}
\widehat\mu_t^{\mathrm{EAM}}
=\sum_{i=1}^{N_t}q_i\,\delta_{\mathbf h_{t,i}^{g}},
\qquad
\widehat\mu_t^{\mathrm{unif}}
=\frac{1}{N_t}\sum_{i=1}^{N_t}\delta_{\mathbf h_{t,i}^{g}}.
\label{eq:tilted-target-measure}
\end{equation}
Here $\delta_{\mathbf h}$ denotes a point mass at embedding $\mathbf h$.
Reliability controls repeated participation, while the positive floor gives
every target node a nonzero probability of contributing to alignment.

At repetition $r$, we draw $m$ source nodes uniformly and $m$ target nodes
from $\mathbf q$, both with replacement. Let
$\mathbf s_{r,a}=\operatorname{sg}[\mathbf h_{s,I_{r,a}^s}^g]$ and
$\mathbf t_{r,a}=\mathbf h_{t,I_{r,a}^t}^g$ denote the $a$th sampled source and
target embeddings, respectively. With repetition-specific kernel $k_r$, EAM
averages their discrepancies over $R$ draws:
\begin{equation}
\mathcal L_{\mathrm{EAM}}
=\frac{1}{Rm^2}\sum_{r=1}^{R}\sum_{a,b=1}^{m}\big[
k_r(\mathbf s_{r,a},\mathbf s_{r,b})
+k_r(\mathbf t_{r,a},\mathbf t_{r,b})
-2k_r(\mathbf s_{r,a},\mathbf t_{r,b})\big].
\label{eq:eam-objective}
\end{equation}
The source--source and target--target sums retain pairs with $a=b$, matching the biased kernel
estimator used in our implementation \citep{gretton2012kernel}. EAM changes the
target samples entering this estimator while leaving its kernel form unchanged;
categorical resampling is a Monte Carlo realization of
$\widehat\mu_t^{\mathrm{EAM}}$, not a continuous weighted-MMD surrogate.
Source embeddings are detached, whereas target embeddings retain gradients.
For each repetition, $k_r$ sums five RBF kernels with bandwidths
$b_r2^{j-2}$, $j=0,\ldots,4$. Here $b_r$ is the detached mean squared distance
between distinct pooled sample positions, clamped away from zero.
Both variants use $m=\min(1000,N_s,N_t)$ samples and $R=5$ repetitions in
every transfer task.

\subsection{Independent Training and Post-training Fusion}
\label{sec:independent-fusion}

Let $\boldsymbol\theta_g$ and $\boldsymbol\theta_\ell$ denote the disjoint
parameter sets of the graph-aware and graph-free local experts. For
$b\in\{g,\ell\}$, source supervision is measured by
\begin{equation}
\mathcal L_{\mathrm{src}}^b
=-\frac{1}{N_s}\sum_{i=1}^{N_s}
\log [\boldsymbol\pi_{s,i}^b]_{y_{s,i}}.
\label{eq:source-classification}
\end{equation}
For $\lambda\geq0$, the two parameter sets are optimized with
\begin{equation}
\min_{\boldsymbol\theta_g}
\big[\mathcal L_{\mathrm{src}}^g(\mathcal G_s;\boldsymbol\theta_g)
+\lambda\mathcal L_{\mathrm{EAM}}
(\mathcal G_s,\mathcal G_t;\boldsymbol\theta_g)\big],
\qquad
\min_{\boldsymbol\theta_\ell}
\mathcal L_{\mathrm{src}}^\ell(\mathcal G_s;\boldsymbol\theta_\ell).
\label{eq:independent-optimization}
\end{equation}
Here $\lambda$ balances source classification and graph-expert alignment; the
graph-free local expert receives source supervision alone. The graph objective
may use both $\mathcal G_s$ and $\mathcal G_t$, but target labels occur in
neither objective. Under the gradient convention in Section~\ref{sec:eam}, EAM
reaches $\boldsymbol\theta_g$ through sampled target embeddings while its source
embeddings are fixed anchors. No term sends an EAM gradient to
$\boldsymbol\theta_\ell$. Thus, independent training means disjoint parameters,
optimizers, and objectives---not merely two classifier heads attached to one
adapted representation.

For transfer $\tau=(s\!\to\!t)$, let $\boldsymbol\pi_{\tau,i}^g$ and
$\boldsymbol\pi_{\tau,i}^\ell$ denote the probabilities returned for target
node $i$ by the fitted experts. After training, a task-level coefficient
$\alpha_\tau\in[0,1]$ forms
\begin{equation}
\boldsymbol\pi_{\tau,i}
=(1-\alpha_\tau)\boldsymbol\pi_{\tau,i}^g
+\alpha_\tau\boldsymbol\pi_{\tau,i}^\ell,
\qquad
\widehat y_{\tau,i}=\arg\max_k[\boldsymbol\pi_{\tau,i}]_k.
\label{eq:probability-fusion}
\end{equation}
Since both expert outputs lie in the probability simplex, Eq.~\ref{eq:probability-fusion}
is also a valid probability vector for every $\alpha_\tau\in[0,1]$. The same
coefficient is shared across target nodes, so fusion introduces neither a
node-wise router nor a pseudo-label objective. Fusion is performed after expert
training and introduces no additional gradient update. We select the task
coefficient from a shared finite grid under the evaluation protocol in
Section~\ref{sec:experimental-setup} for all 16 transfer directions.

%% file: arxiv/sections/04_experiments.tex
\section{Experiments}
\label{sec:experiments}

We evaluate \method{} on 16 transfers to assess adaptation performance and
identify the contributions of its two prediction paths and alignment mechanism.
The experiments connect expert-level evidence to matched controls, topology
interventions, and fusion sensitivity through five research questions:
\begin{itemize}[leftmargin=*,itemsep=0pt,parsep=0pt]
    \item \textbf{RQ1.} \emph{How effective is \method{} across graph domains?}
    Section~\ref{sec:overall-performance} compares its adaptation performance
    with representative GDA methods on 16 transfers across four benchmark families.
    \item \textbf{RQ2.} \emph{Do the experts provide complementary evidence?}
    Sections~\ref{sec:predictive-complementarity} and~\ref{sec:topology-stress}
    analyze expert-specific correctness and controlled topology corruption
    to examine the value of graph-free predictions.
    \item \textbf{RQ3.} \emph{Does entropy-aware sampling improve alignment?}
    Section~\ref{sec:eam-results} compares EAM with matched uniform sampling
    to isolate the contribution of entropy-guided target participation.
    \item \textbf{RQ4.} \emph{Can capacity or generic ensembling explain the gains?}
    Section~\ref{sec:component-attribution} compares the heterogeneous pair
    with parameter-matched Graph-only and Graph+Graph controls.
    \item \textbf{RQ5.} \emph{How sensitive and costly is task-level fusion?}
    Section~\ref{sec:sensitivity-cost} examines responses to the fusion coefficient
    and measures the local expert's training, inference, and memory overhead.
\end{itemize}

\subsection{Experimental Setup}
\label{sec:experimental-setup}

\paragraph{Datasets and metrics.}
We evaluate ten processed graphs and 16 directed transfer tasks adopted by
recent graph domain adaptation studies
\citep{liu2024a2gnn,chen2026adalign}.
The Citation family contains ACMv9 (A), Citationv1 (C), and DBLPv7 (D);
the Airport family contains Brazil (B), Europe (E), and USA (U);
the Blog family contains Blog1 (B1) and Blog2 (B2);
and the Twitch family contains German (DE) and English (EN).
Citation and Airport contribute six directed transfers each, whereas Blog
and Twitch contribute two transfers each.
All methods read the same processed tensors, preserving the supplied node
features and edge representation for each transfer direction.

Macro-F1 is the primary evaluation metric.
We additionally report Micro-F1 at the checkpoint selected according to
Macro-F1, denoted by Micro@Macro.
Family-level scores are equal-task averages within each graph family, and
the Full-16 score assigns equal weight to each transfer task.
Neither aggregate is weighted by the number of nodes in a graph.

\paragraph{Baselines.}
We compare against source-only GCN \citep{kipf2017gcn},
alignment-based
UDA-GCN \citep{wu2020udagcn} and GRADE \citep{wu2023grade};
graph-shift-aware
PairAlign \citep{liu2024pairalign},
GraphAlign \citep{huang2024graphalign},
A2GNN \citep{liu2024a2gnn},
TDSS \citep{chen2025tdss},
DGSDA \citep{yang2025dgsda},
GAA \citep{fang2025attrgda},
and HGDA \citep{fang2025hgda};
and recent adaptation methods
ADAlign \citep{chen2026adalign},
DiffGDA \citep{chen2026diffgda},
and DFT \citep{tai2026dft}.
ADAlign and DiffGDA represent recent state-of-the-art approaches to our
knowledge. Graph-only serves as our backbone control for measuring gains from
the full method.

\paragraph{Implementation and evaluation protocol.}
Following recent GDA evaluation practice~\citep{chen2026adalign}, all locally
reproduced methods are trained for 150 epochs and evaluated over the same five
runs using a unified evaluator; we report the mean and sample standard
deviation across runs. \method{} uses Adam, graph/local hidden widths of
128/64, one graph feature layer, ReLU, and no dropout. Only the graph learning
rate, propagation pair, and alignment weight vary during training; all other
architecture, regularization, and estimator settings remain fixed.
The graph-aware and graph-free local experts are optimized independently, and
their output probabilities are combined after training through a task-level
constant mixture. Each task evaluates 96 family-level training bundles and
replays fusion over a shared coefficient grid. Appendix~\ref{app:protocol}
lists the family-level search spaces and globally fixed implementation settings.
The server has dual Intel Xeon E5-2680 v4 CPUs and 48\,GB NVIDIA GeForce
RTX 4090 D GPUs; each training run uses one GPU.

\paragraph{Statistical analysis.}
Controlled contrasts pair task and run, reporting task-level win/tie/loss counts.
Expert and capacity comparisons use 10,000 bootstrap replicates, resampling
tasks and then paired runs within each task. The 95\% intervals span the
2.5th--97.5th percentiles of the aggregate paired-difference distribution,
retaining equal task weights in each replicate.

%% file: arxiv/sections/05_results.tex
\begingroup
\setlength{\parskip}{4pt plus 2pt}
\clubpenalty=500
\widowpenalty=500
\makeatletter
\newcommand{\fixedresultsubsection}{%
  \@startsection{subsection}{2}{\z@}{-1.8ex}{0.8ex}%
  {\normalsize\sc\raggedright}}
\makeatother

\fixedresultsubsection{Overall Adaptation Performance}
\label{sec:overall-performance}

\textbf{Adaptation versus source-only learning.}
First, UDA-GCN and GRADE outperform source-only GCN on all 16 transfers
in Table~\ref{tab:main-results}. On B1$\rightarrow$B2, Macro-F1 increases
from $21.96$ with GCN to $30.94$ and $40.56$, respectively.
These consistent gains highlight the importance of addressing cross-domain
discrepancies: learning a graph classifier from source supervision alone does
not ensure that its representations remain predictive in the target domain.

\begingroup
\noindent\begin{minipage}{\textwidth}
    \centering
    \captionof{table}{
        \textbf{Macro-F1 on 16 graph-domain transfers.}
        Mean (\%) and sample standard deviation over five runs; best and
        second-best results are bold and underlined.
    }
    \label{tab:main-results}
    \resizebox{\textwidth}{!}{%
        \input{arxiv/tables/main_comparison}
    }
\end{minipage}
\endgroup

\textbf{Progress in graph adaptation.}
Second, recent methods further improve representative transfer directions.
Compared with UDA-GCN, ADAlign's adaptive alignment raises C$\rightarrow$D
from $75.17$ to $76.59$, while DGSDA's separation of attribute and topology
adaptation raises D$\rightarrow$C from $72.23$ to $81.49$.
These improvements suggest that effective transfer depends not only on reducing
distribution discrepancy, but also on how attributes and relational information
are used. This motivates examining whether complementary information should
remain separately accessible at prediction time, rather than deriving every
target decision from one adapted graph representation.

\textbf{Complementary prediction paths.}
Finally, \method{} achieves $67.67 \pm 0.04$ Full-16 Macro-F1 and ranks first
on 15 transfers, including all six Airport and both Blog and Twitch directions.
It exceeds the strongest competing results by $4.74$ points on U$\rightarrow$E
and $4.27$ on E$\rightarrow$B. This success is attributed to preserving two
complementary prediction paths: the graph-aware expert captures relational
information, while the graph-free local expert retains attribute evidence
without neighborhood aggregation. Constant probability fusion combines these
predictions, and entropy-aware sampling prioritizes confident target evidence
during graph alignment. The following controls examine the contributions of
expert complementarity and sampling importance separately.

\fixedresultsubsection{Complementary Predictive Evidence}
\label{sec:predictive-complementarity}

Figure~\ref{fig:complementarity} partitions post-training target correctness
for Local or Graph~2 against the same graph-aware expert and compares Graph-only
with fused endpoints on representative transfers.

\suppressfloats[t]
\begin{figure}[tbp]
    \centering
    \includegraphics[width=\textwidth]{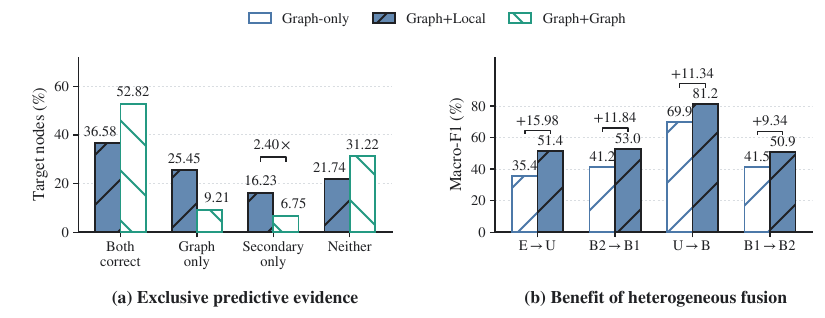}
    \caption{
        \textbf{Heterogeneous predictive complementarity.}
        (a) Full-16 correctness: Local supplies $2.40\times$ the exclusive
        correct predictions of Graph~2 against the same graph expert.
        (b) Representative transfers: Graph-only versus Graph+Local,
        with bracketed fusion gains in Macro-F1 points.
    }
    \label{fig:complementarity}
\end{figure}

Fusion exceeds the better single expert by $1.98$ points ($95\%$ CI
$[0.67,3.67]$), with all five run-level aggregates positive. The local expert
is exclusively correct on 16.23\% of target nodes, more than twice the 6.75\%
supplied by a second graph expert. Fusion corrects graph errors on 10.87\% of
target nodes and introduces errors on 4.45\%. The resulting gain reflects useful
local-exclusive predictions on nodes where the graph-aware expert is incorrect.

Graph+Graph instead yields $-0.82\%$ net repairs, indicating that independent
initialization alone need not provide useful corrections to the graph expert's errors.

The corresponding fusion gains are $0.05$, $8.21$, $10.59$, and $1.42$
points on Citation, Airport, Blog, and Twitch. The larger Airport and Blog
responses align with greater local-exclusive evidence. We examine this
relationship directly through the degree-preserving topology intervention in
Section~\ref{sec:topology-stress}.

\fixedresultsubsection{Entropy-Aware Alignment}
\label{sec:eam-results}

We test whether predictive confidence informs target participation in marginal
alignment. EAM changes the empirical target measure, not the discrepancy objective:
lower-entropy predictions recur more often across draws, while the sampling floor
retains every node's support. Thus, node participation changes while the
discrepancy criterion remains fixed.

\noindent\begin{minipage}[t]{0.40\textwidth}
    \vspace{0pt}
    \centering
    \includegraphics[width=\linewidth]{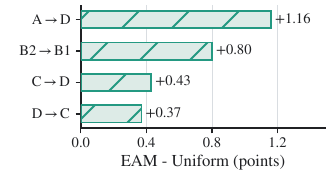}
    \captionof{figure}{
        \textbf{Matched entropy-aware resampling.}
        EAM gains on representative transfers; Appendix~\ref{app:eam-control}
        reports all 16 matched task comparisons.
    }
    \label{fig:eam-representative}
\end{minipage}\hfill
\begin{minipage}[t]{0.57\textwidth}
    \vspace{0pt}
    \textbf{Matched comparison.}
    The Uniform variant assigns every target node the same sampling probability
    while leaving the experts, sample count, replacement semantics,
    multi-kernel MMD, bandwidth rule, optimizer, and source/target gradient
    routes unchanged throughout training.

    \smallskip
    \textbf{Alignment contribution.}
    EAM improves the fused endpoint by $0.24$ Macro-F1 points on average,
    with family effects of $+0.35$, $+0.06$, $+0.59$, and $+0.08$ on Citation,
    Airport, Blog, and Twitch. Figure~\ref{fig:eam-representative} shows
    gains of $1.16$, $0.80$, $0.43$, and $0.37$ points on A$\rightarrow$D,
    B2$\rightarrow$B1, C$\rightarrow$D, and D$\rightarrow$C.
    Entropy-guided participation strengthens these alignment estimates by
    concentrating repeated draws on confident target evidence from the graph expert.
\end{minipage}

\fixedresultsubsection{Controlled Topology-Shift Stress Test}
\label{sec:topology-stress}

We progressively rewire target edges on A$\rightarrow$C, U$\rightarrow$B,
B1$\rightarrow$B2, and EN$\rightarrow$DE, using six severity levels and five
runs per transfer. Rewiring preserves edge count and every node's directed
degree while leaving attributes and local predictions unchanged.
Figure~\ref{fig:topology-shift} tracks how this controlled neighborhood
corruption changes graph-expert performance and local-exclusive evidence.

Homophily loss tracks graph damage ($\rho=0.943$), which in turn tracks
local-exclusive evidence ($\rho=0.973$). At maximum corruption, Graph-only
drops by 0.270 while the local-exclusive rate rises by 0.141. Under this
matched intervention, the trajectories isolate the increasing relative value
of topology-free evidence. A$\rightarrow$C responds most strongly and
U$\rightarrow$B least, showing that the value of topology-free evidence
depends on the transfer's response to neighborhood corruption.

\begingroup
\begin{figure}[H]
    \centering
    \includegraphics[width=0.90\linewidth]{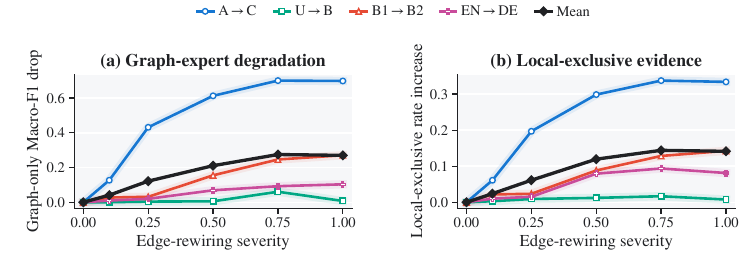}
    \caption{\textbf{Topology corruption exposes heterogeneous evidence.}
    Graph-expert degradation and local-exclusive correctness under
    degree-preserving edge rewiring; colored curves denote representative
    transfers, with their equal-task mean shown as the black reference curve.}
    \label{fig:topology-shift}
\end{figure}
\endgroup

\fixedresultsubsection{Controlled Attribution}
\label{sec:component-attribution}
\label{sec:capacity-controls}
\label{sec:sensitivity-cost}

We compare with a graph model widened to match Full within five parameters
and Graph+Graph, which replaces Local with an independently initialized graph
expert. The graph pair preserves ensembling while giving both experts the
same inputs. All reviewer
controls use the frozen task configuration of the full method, the same
training and checkpoint budgets, and identical evaluation procedures. Only
the expert identity or graph-model capacity is changed. Graph+Local gains
$5.13$ points over the capacity control (14 task wins), $4.84$ over Graph+Graph
(12 wins, four ties), and $1.98$ over the better single expert
(Table~\ref{tab:core-controls}). Together with Figure~\ref{fig:complementarity}(a),
these controls support the value of heterogeneous information access beyond
added capacity and generic averaging. The local expert retains an
attribute-only prediction path alongside the adapted graph representation,
allowing task-level probability fusion to exploit complementary predictions
from independently trained experts.

\noindent\begin{minipage}[t]{0.49\textwidth}
    \vspace{0pt}
    \raggedright
    \textbf{Fusion response.}
    Frozen-logit responses peak at an interior coefficient. E$\rightarrow$U and
    B2$\rightarrow$B1 retain broad beneficial regions; U$\rightarrow$B favors
    graph-dominant fusion.
    \textbf{Efficiency.} Without routing or extra graph propagation, the local
    expert adds about $2\%$ training time and $9.38\%$ inference latency, with
    unchanged peak memory on A$\rightarrow$C (RTX 4090 D). These final-model
    costs exclude the one-time hyperparameter search.

    \smallskip
    \centering
    \captionof{table}{\textbf{Paired gains over capacity and ensemble controls}
    (percentage points; 95\% CIs).}
    \label{tab:core-controls}
    \scriptsize
    \setlength{\tabcolsep}{1pt}
    \input{arxiv/tables/core_controls}

\end{minipage}\hfill
\begin{minipage}[t]{0.48\textwidth}
    \vspace{0pt}
    \centering
    \includegraphics[width=\linewidth]{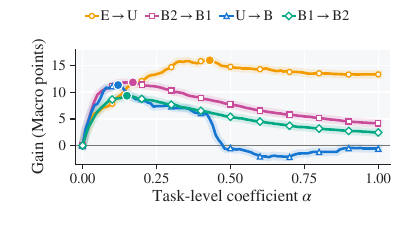}
    \captionof{figure}{
        \textbf{Task-level fusion response.}
        Curves replay frozen expert predictions; filled markers denote observed
        maxima. Soft under-strokes emphasize the response trajectories.
    }
    \label{fig:fusion-response}

\end{minipage}
\endgroup

%% file: arxiv/tables/main_comparison.tex
\begin{tabular}{lcccccccc}
\toprule
Method & A$\to$C & A$\to$D & C$\to$A & C$\to$D & D$\to$A & D$\to$C & B1$\to$B2 & B2$\to$B1 \\
\midrule
\multicolumn{9}{l}{\emph{Source-only}} \\
GCN \venue{ICLR'17}
& \mstd{66.97}{1.62} & \mstd{61.38}{1.10} & \mstd{66.82}{0.85} & \mstd{66.97}{0.70} & \mstd{57.34}{1.32} & \mstd{65.60}{2.53} & \mstd{21.96}{1.88} & \mstd{23.05}{2.04} \\
\midrule
\multicolumn{9}{l}{\emph{Graph domain adaptation}} \\
UDAGCN \venue{WWW'20}
& \mstd{77.62}{1.35} & \mstd{71.59}{2.31} & \mstd{73.56}{1.56} & \mstd{75.17}{2.65} & \mstd{62.17}{1.98} & \mstd{72.23}{2.30} & \mstd{30.94}{1.13} & \mstd{31.34}{1.15} \\
GRADE \venue{AAAI'23}
& \mstd{68.06}{2.29} & \mstd{63.94}{2.22} & \mstd{68.46}{0.40} & \mstd{69.77}{1.17} & \mstd{63.76}{0.61} & \mstd{65.74}{0.95} & \mstd{40.56}{1.37} & \mstd{42.41}{2.76} \\
PairAlign \venue{ICML'24}
& \mstd{57.09}{1.61} & \mstd{55.25}{2.05} & \mstd{52.20}{2.50} & \mstd{56.73}{3.25} & \mstd{49.34}{1.36} & \mstd{52.31}{2.27} & \mstd{40.99}{2.07} & \mstd{43.73}{0.68} \\
GraphAlign \venue{KDD'24}
& \mstd{67.07}{1.29} & \mstd{63.05}{1.79} & \mstd{63.57}{0.90} & \mstd{65.99}{1.46} & \mstd{59.02}{0.97} & \mstd{62.15}{0.93} & \mstd{35.85}{1.85} & \mstd{39.60}{1.23} \\
A2GNN-MMD \venue{AAAI'24}
& \mstd{78.96}{0.42} & \mstd{72.66}{0.46} & \mstd{75.17}{0.28} & \mstd{74.70}{0.73} & \mstd{73.96}{0.61} & \mstd{78.11}{0.98} & \mstd{44.15}{2.30} & \mstd{43.77}{0.83} \\
TDSS-RW \venue{AAAI'25}
& \mstd{79.67}{0.19} & \mstd{73.71}{0.69} & \mstd{\underline{76.16}}{0.42} & \mstd{72.72}{0.42} & \mstd{74.70}{0.46} & \mstd{76.43}{0.27} & \mstd{42.81}{1.15} & \mstd{45.11}{1.28} \\
GAA \venue{ICLR'25}
& \mstd{71.97}{0.66} & \mstd{64.84}{1.36} & \mstd{69.32}{0.35} & \mstd{69.88}{0.88} & \mstd{63.00}{0.57} & \mstd{68.73}{1.68} & \mstd{44.61}{2.34} & \mstd{40.79}{1.39} \\
DGSDA \venue{ICML'25}
& \mstd{\underline{80.54}}{0.39} & \mstd{74.13}{1.01} & \mstd{76.04}{0.19} & \mstd{75.89}{0.22} & \mstd{\underline{74.90}}{0.36} & \mstd{\textbf{81.49}}{0.50} & \mstd{36.33}{1.70} & \mstd{39.67}{2.38} \\
HGDA \venue{ICML'25}
& \mstd{71.58}{0.96} & \mstd{65.00}{1.03} & \mstd{67.59}{0.77} & \mstd{69.09}{0.97} & \mstd{66.21}{0.83} & \mstd{70.86}{0.99} & \mstd{43.32}{0.40} & \mstd{43.64}{0.36} \\
ADAlign \venue{ICLR'26}
& \mstd{79.03}{0.38} & \mstd{\underline{75.14}}{1.55} & \mstd{75.25}{0.37} & \mstd{\underline{76.59}}{1.01} & \mstd{72.15}{1.31} & \mstd{75.40}{1.70} & \mstd{40.81}{2.52} & \mstd{40.68}{1.09} \\
DiffGDA \venue{ICLR'26}
&  \mstd{78.83}{1.09} & \mstd{72.67}{0.89} &\mstd{ 74.88}{0.86} & \mstd{73.33}{1.21} & \mstd{71.26}{2.70} &\mstd{75.45}{1.44} & \mstd{42.69}{2.07} & \mstd{40.15}{2.88} \\
DFT \venue{KDD'26}
& \mstd{72.59}{1.83} & \mstd{69.69}{1.02} & \mstd{68.06}{0.73} & \mstd{70.90}{1.92} & \mstd{65.31}{2.71} & \mstd{75.35}{1.92} & \mstd{\underline{47.25}}{8.35} & \mstd{\underline{51.75}}{4.24} \\
\midrule
\rowcolor{oursrow}
\textbf{\method{} (Ours)}
& \mstd{\textbf{81.93}}{0.28} & \mstd{\textbf{76.18}}{1.48} & \mstd{\textbf{77.34}}{0.19} & \mstd{\textbf{78.08}}{0.41} & \mstd{\textbf{75.48}}{0.15} & \mstd{\underline{81.24}}{0.54} & \mstd{\textbf{50.88}}{1.73} & \mstd{\textbf{53.01}}{0.61} \\
\midrule
Method & U$\to$B & U$\to$E & B$\to$U & B$\to$E & E$\to$U & E$\to$B & DE$\to$EN & EN$\to$DE \\
\midrule
\multicolumn{9}{l}{\emph{Source-only}} \\
GCN \venue{ICLR'17}
& \mstd{45.06}{3.21} & \mstd{34.17}{2.12} & \mstd{29.41}{0.32} & \mstd{25.27}{0.86} & \mstd{30.30}{0.22} & \mstd{32.66}{1.84} & \mstd{56.23}{0.32} & \mstd{57.31}{1.86} \\
\midrule
\multicolumn{9}{l}{\emph{Graph domain adaptation}} \\
UDAGCN \venue{WWW'20}
& \mstd{60.82}{22.45} & \mstd{41.28}{1.77} & \mstd{39.41}{2.55} & \mstd{50.03}{0.80} & \mstd{40.35}{1.06} & \mstd{63.89}{1.69} & \mstd{57.91}{0.47} & \mstd{58.33}{1.54} \\
GRADE \venue{AAAI'23}
& \mstd{61.48}{4.34} & \mstd{47.61}{1.00} & \mstd{38.50}{3.07} & \mstd{\underline{55.45}}{1.90} & \mstd{43.96}{1.39} & \mstd{70.38}{1.48} & \mstd{58.72}{0.06} & \mstd{62.12}{0.17} \\
PairAlign \venue{ICML'24}
& \mstd{70.03}{1.82} & \mstd{39.29}{1.42} & \mstd{45.35}{4.19} & \mstd{38.88}{3.89} & \mstd{40.50}{2.84} & \mstd{48.45}{3.02} & \mstd{60.11}{0.37} & \mstd{\underline{63.72}}{0.16} \\
GraphAlign \venue{KDD'24}
& \mstd{63.69}{2.27} & \mstd{53.24}{1.20} & \mstd{46.21}{2.86} & \mstd{54.18}{1.59} & \mstd{48.83}{2.61} & \mstd{68.54}{1.02} & \mstd{54.27}{1.01} & \mstd{55.07}{2.61} \\
A2GNN-MMD \venue{AAAI'24}
& \mstd{60.36}{1.56} & \mstd{46.52}{1.03} & \mstd{43.10}{1.81} & \mstd{48.12}{1.06} & \mstd{40.28}{5.22} & \mstd{64.11}{6.61} & \mstd{56.71}{0.48} & \mstd{56.15}{1.49} \\
TDSS-RW \venue{AAAI'25}
& \mstd{74.56}{0.60} & \mstd{41.89}{0.28} & \mstd{53.96}{1.68} & \mstd{43.26}{3.02} & \mstd{48.23}{0.07} & \mstd{59.29}{0.36} & \mstd{53.62}{0.91} & \mstd{40.07}{1.67} \\
GAA \venue{ICLR'25}
& \mstd{70.42}{1.34} & \mstd{\underline{54.02}}{3.25} & \mstd{55.66}{0.73} & \mstd{53.40}{2.51} & \mstd{{50.68}}{1.20} & \mstd{\underline{71.48}}{1.88} & \mstd{46.11}{1.75} & \mstd{46.45}{1.02} \\
DGSDA \venue{ICML'25}
& \mstd{63.37}{1.30} & \mstd{{52.87}}{1.20} & \mstd{\underline{56.17}}{0.35} & \mstd{51.87}{1.32} & \mstd{\underline{50.75}}{0.39} & \mstd{70.36}{1.12} & \mstd{59.92}{0.12} & \mstd{61.49}{0.18} \\
HGDA \venue{ICML'25}
& \mstd{71.83}{22.93} & \mstd{43.56}{3.34} & \mstd{52.91}{0.95} & \mstd{50.49}{2.00} & \mstd{48.12}{0.71} & \mstd{66.11}{1.82} & \mstd{59.52}{0.34} & \mstd{60.84}{0.64} \\
ADAlign \venue{ICLR'26}
& \mstd{75.56}{2.31} & \mstd{52.23}{1.29} & \mstd{47.86}{1.36} & \mstd{55.13}{1.25} & \mstd{50.27}{0.79} & \mstd{67.87}{1.27} & \mstd{59.33}{0.10} & \mstd{62.94}{0.40} \\
DiffGDA \venue{ICLR'26}
& \mstd{75.73}{1.41} & \mstd{47.07}{1.34} & \mstd{45.64}{0.94} & \mstd{53.09}{0.82} & \mstd{41.07}{1.38} & \mstd{62.70}{0.97} & \mstd{56.02}{0.23} & \mstd{57.65}{1.34} \\
DFT \venue{KDD'26}
& \mstd{\underline{76.62}}{0.46} & \mstd{50.11}{1.15} & \mstd{53.58}{1.16} & \mstd{53.32}{1.78} & \mstd{49.52}{1.09} & \mstd{69.10}{1.04} & \mstd{\underline{60.55}}{0.60} & \mstd{63.63}{1.21} \\
\midrule
\rowcolor{oursrow}
\textbf{\method{} (Ours)}
& \mstd{\textbf{81.20}}{0.35} & \mstd{\textbf{58.76}}{0.60} & \mstd{\textbf{56.90}}{0.32} & \mstd{\textbf{59.76}}{0.14} & \mstd{\textbf{51.41}}{1.08} & \mstd{\textbf{75.75}}{1.30} & \mstd{\textbf{60.58}}{0.09} & \mstd{\textbf{64.22}}{0.41} \\
\bottomrule
\end{tabular}

%% file: arxiv/tables/core_controls.tex
\begin{tabular}{@{}lrrr@{}}
\toprule
Control & $\Delta$ & 95\% CI & W/T/L \\
\midrule
Capacity-matched graph & $+5.13$ & $[1.93,9.05]$ & 14/0/2 \\
Graph+Graph & $+4.84$ & $[1.72,8.60]$ & 12/4/0 \\
Best expert & $+1.98$ & $[0.67,3.67]$ & 12/4/0 \\
\bottomrule
\end{tabular}

%% file: arxiv/sections/06_discussion.tex
\section{Conclusion}
\label{sec:conclusion}

\method{} complements representation alignment with independently trained
graph-aware and graph-free local experts, entropy-aware exact sampling, and
task-level probability fusion. It achieves the best Macro-F1 on 15 of 16
transfers and improves over the stronger expert by 1.98 points on average.
Capacity and ensemble controls support heterogeneous information access as
the source of these gains.
The current study focuses on static graphs with a single labeled source domain.
Future work will support
evolving graphs through incremental expert updates and refreshed alignment
samples, and multi-source adaptation through source-specific graph experts and
source-level probability fusion.\par

%% file: arxiv/sections/07_statements.tex
\section*{AI Use Statement}

Generative AI tools were used to assist with literature organization,
research brainstorming, code review, result-consistency checking,
scientific-figure preparation, and manuscript editing.
The authors designed the methods and experimental protocols, executed all
experiments, verified the reported measurements against the underlying
artifacts, checked the cited literature, and made all final scientific and
editorial decisions.
No generative AI tool was used to generate benchmark data, target labels, or
experimental measurements.
The authors take full responsibility for the contents of this paper.

\section*{Reproducibility Statement}

Section~\ref{sec:method} specifies the model architecture, optimization
objectives, gradient paths, entropy-aware sampling procedure, and
post-training fusion rule.
Section~\ref{sec:experiments} describes the datasets, evaluation metrics,
baseline comparisons, repeated runs, and statistical analyses.
The appendix provides estimator details, preprocessing conventions,
additional results, matched-control specifications, parameter analyses,
and computational-cost measurements.
The supplementary material provides the implementation and scripts used to
reproduce the reported tables and figures.

%% file: arxiv/sections/appendix.tex
\raggedbottom
\clubpenalty=10000
\widowpenalty=10000
\displaywidowpenalty=10000
\setlength{\textfloatsep}{9pt plus 2pt minus 2pt}
\setlength{\floatsep}{8pt plus 2pt minus 2pt}
\setlength{\intextsep}{8pt plus 2pt minus 2pt}
\renewcommand{\topfraction}{0.92}
\renewcommand{\bottomfraction}{0.85}
\renewcommand{\textfraction}{0.08}
\renewcommand{\floatpagefraction}{0.82}
\setcounter{topnumber}{3}
\setcounter{bottomnumber}{2}
\setcounter{totalnumber}{5}
\section{Theoretical Results}
\label{app:theory}

\subsection{Prediction-space complementarity}
\label{app:complementarity-theory}

\paragraph{Proof of Proposition~\ref{prop:residual-fusion}.}
For any input $S$, the conditional distribution
$q^\star(\cdot\mid S)=P_t(Y=\cdot\mid S)$ minimizes expected log-loss, so
$\mathcal R_{\log}^{\star}(S)=H_t(Y\mid S)$. The chain rule expresses the
reduction in optimal log-risk as conditional mutual information:
\begin{equation}
\mathcal R_{\log}^{\star}(Z^g)-
\mathcal R_{\log}^{\star}(Z^g,X)
=H_t(Y\mid Z^g)-H_t(Y\mid Z^g,X)
=I_t(Y;X\mid Z^g).
\label{eq:residual-information-proof}
\end{equation}
It vanishes exactly when $Y$ and $X$ are conditionally independent given $Z^g$.

Let $Y$ be a target label, $\mathbf e_Y$ its one-hot vector, and
$\boldsymbol\pi^g,\boldsymbol\pi^\ell\in\Delta^{K-1}$ the two predictive
distributions for the same target node. For
$\boldsymbol\pi^\alpha=(1-\alpha)\boldsymbol\pi^g+
\alpha\boldsymbol\pi^\ell$ with $\alpha\in[0,1]$, define the Brier risk as
$\mathcal R(\boldsymbol\pi)=
\mathbb E\lVert\mathbf e_Y-\boldsymbol\pi\rVert_2^2$, where the expectation is
over the target distribution. Expanding the squared norm gives
\begin{equation}
\begin{aligned}
\mathcal R(\boldsymbol\pi^\alpha)
={}&(1-\alpha)\mathcal R(\boldsymbol\pi^g)
+\alpha\mathcal R(\boldsymbol\pi^\ell)\\
&-\alpha(1-\alpha)
\mathbb E\lVert\boldsymbol\pi^g-\boldsymbol\pi^\ell\rVert_2^2.
\end{aligned}
\label{eq:brier-decomposition}
\end{equation}
Indeed, applying
$\lVert(1-\alpha)\mathbf a+\alpha\mathbf b\rVert_2^2
=(1-\alpha)\lVert\mathbf a\rVert_2^2
+\alpha\lVert\mathbf b\rVert_2^2
-\alpha(1-\alpha)\lVert\mathbf a-\mathbf b\rVert_2^2$
to $\mathbf a=\mathbf e_Y-\boldsymbol\pi^g$ and
$\mathbf b=\mathbf e_Y-\boldsymbol\pi^\ell$, then taking expectations,
proves Eq.~\ref{eq:brier-decomposition}. For $\alpha>0$, fusion improves the
graph-aware expert whenever
\begin{equation}
(1-\alpha)\mathbb E\lVert\boldsymbol\pi^g-\boldsymbol\pi^\ell\rVert_2^2
>\mathcal R(\boldsymbol\pi^\ell)-\mathcal R(\boldsymbol\pi^g),
\label{eq:complementarity-condition}
\end{equation}
obtained by subtracting $\mathcal R(\boldsymbol\pi^g)$ from both sides of
Eq.~\ref{eq:brier-decomposition}. The disagreement term measures predictive
diversity, while the condition identifies when this diversity is large enough
to offset the local expert's risk gap.
To prove the second part of Proposition~\ref{prop:residual-fusion}, set
$\Delta=\mathcal R(\boldsymbol\pi^\ell)-
\mathcal R(\boldsymbol\pi^g)\geq0$ and
$D=\mathbb E\lVert\boldsymbol\pi^g-
\boldsymbol\pi^\ell\rVert_2^2$. Equation~\ref{eq:brier-decomposition} yields
\begin{equation}
\mathcal R(\boldsymbol\pi^\alpha)-
\mathcal R(\boldsymbol\pi^g)
=\alpha(\Delta-D)+\alpha^2D.
\label{eq:fusion-quadratic}
\end{equation}
If $D\leq\Delta$, the right-hand side is nonnegative for every
$\alpha\in[0,1]$. If $D>\Delta$, its minimizer on $[0,1]$ is
$\alpha^\star=(D-\Delta)/(2D)\in(0,1/2]$, and substitution gives the strict
risk reduction in Eq.~\ref{eq:optimal-fusion-gain}.

Together, conditional information identifies attribute signal beyond the
graph-aware state, while the Brier condition determines whether the trained
graph-free local expert converts that signal into a lower-risk mixture. The
correctness decomposition and matched capacity and ensemble controls evaluate
how these conditions relate to the predictions of the trained experts.

\section{Reproduction Details}
\label{app:reproduction}
\label{app:algorithm}

\subsection{Training and inference}

Each task and run uses the following independent-training and fusion procedure:
\begin{enumerate}
    \item Initialize the two experts with disjoint parameters and separate optimizers.
    \item At each epoch, compute source and target graph embeddings, form
    detached entropy-dependent sampling probabilities, and update the graph
    parameters using $\mathcal L_{\mathrm{src}}^g+
    \lambda\mathcal L_{\mathrm{EAM}}$.
    \item Update the local MLP once on source cross-entropy with its own optimizer.
    \item Store both experts' target logits at each epoch for probability-fusion replay.
\end{enumerate}
The optimizers have disjoint parameter groups. EAM gradients remain within the
graph parameter block, and fusion operates on stored predictions after training.

\subsection{Graph-aware and graph-free local experts}
\label{app:graph-encoder}

For the graph-aware expert, let $\widetilde{\mathbf A}_d$ be obtained from the
processed adjacency matrix $\mathbf A_d$ by inserting a unit self-loop only
where the diagonal entry is zero. Under the convention that
$[\mathbf A_d]_{ij}>0$ sends a message from node $j$ to node $i$, propagation
uses symmetric degree normalization:
\begin{equation}
\mathbf D_d=\operatorname{diag}(\widetilde{\mathbf A}_d\mathbf 1_{N_d}),
\qquad
\mathbf S_d=\mathbf D_d^{-1/2}
\widetilde{\mathbf A}_d\mathbf D_d^{-1/2}.
\end{equation}
The A2GNN encoder in Eq.~\ref{eq:graph-expert} applies $P_d$ propagation steps
before one feature transformation and ReLU, followed by one graph classification
layer. Thus $P_d=0$ removes propagation from the feature extractor but retains
the graph classifier. The graph-free local expert $F_\ell$ is a two-layer ReLU
MLP applied independently to each row of $\mathbf X_d$.
The graph-aware and local-expert hidden widths are 128 and 64, respectively;
both classification heads return normalized class probabilities.

\subsection{Exact EAM implementation}
\label{app:eam}

Let $m=\min(1000,N_s,N_t)$ and $R=5$. For repetition $r$, source indices are
sampled uniformly with replacement and target indices are sampled with
replacement from Eq.~\ref{eq:entropy-sampling}. Sampling weights and graph
probabilities used to construct $\mathbf q$ are detached; sampled target
embeddings retain gradients. We set the source gradient scale to $\gamma=0$,
so sampled source embeddings serve as fixed anchors. Let
$\mathbf S_r=[\mathbf s_{r,1},\ldots,\mathbf s_{r,m}]^\top$ and
$\mathbf T_r=[\mathbf t_{r,1},\ldots,\mathbf t_{r,m}]^\top$ collect the sampled
embeddings, and let
$\mathbf U_r=[\mathbf S_r;\mathbf T_r]\in\mathbb R^{2m\times h_g}$
with $\mathbf u_{r,a}$ denoting its $a$th row. We compute a detached base bandwidth
for each repetition and combine five RBF kernels with geometric bandwidths:
\begin{equation}
\begin{aligned}
\bar b_r
&=\frac{1}{2m(2m-1)}
\sum_{\substack{a,b=1\\a\ne b}}^{2m}
\lVert\mathbf u_{r,a}-\mathbf u_{r,b}\rVert_2^2,
&b_r&=\max\{\operatorname{sg}[\bar b_r],\epsilon_{\mathrm{mach}}\},\\
k_r(\mathbf u,\mathbf v)
&=\sum_{j=0}^{4}\exp\!\left(
-\frac{\lVert\mathbf u-\mathbf v\rVert_2^2}
{b_r\,2^{j-2}}\right).
\end{aligned}
\label{eq:eam-kernel}
\end{equation}
Together with Eq.~\ref{eq:eam-objective}, this specifies the exact biased MMD
V-statistic used in training. The matched Uniform control sets $q_i=1/N_t$ and
executes this same replacement sampler, bandwidth computation, kernel mixture,
and estimator for both source and target domains.

With the fixed floor $\varepsilon=0.8$, the unnormalized sampling weights
lie in $[0.8,1]$, so the sampling probabilities of any two target nodes differ
by a factor of at most $1.25$. EAM therefore adjusts participation continuously
rather than selecting a hard confidence subset. Replacement draws permit a
node to contribute more than once, and averaging repeated estimates exposes
the alignment objective to multiple sampled node sets. The floor controls
relative participation, while the sample cap and repetition count control
the computational budget of the discrepancy estimate.

\subsection{Shared settings and family-level search spaces}
\label{app:protocol}

Tables~\ref{tab:frozen-settings} and~\ref{tab:formal-search-space} list the
fixed settings and family-level search spaces.

\begin{table}[H]
    \centering
    \caption{Shared architecture, optimization, and estimator settings.}
    \label{tab:frozen-settings}
    \small
    \input{arxiv/tables/frozen_settings}
\end{table}

The trained bundle varies the graph learning rate, source/target propagation
pair, and alignment weight. Every task evaluates the same
$3\times8\times4=96$ family-level bundles in
Table~\ref{tab:formal-search-space}.
Thus, the trained search has three task-sensitive dimensions. Architecture
and estimator settings are fixed globally; weight decay is fixed within each
family as listed in Table~\ref{tab:formal-search-space}, not retuned per transfer.
The task-level fusion coefficient is evaluated by replaying the stored expert
predictions. All transfers use the same
coarse candidate set $\alpha_\tau\in\{0.05,0.1,0.4,0.6\}$. Within each
transfer, one fixed coefficient combines the graph-aware and graph-free
probability vectors for all target nodes.

\begin{table}[H]
    \centering
    \caption{Family-level search spaces for the 96 trained bundles per task.}
    \label{tab:formal-search-space}
    \footnotesize
    \setlength{\tabcolsep}{3pt}
    \input{arxiv/tables/formal_search_space}
\end{table}

\subsection{Datasets and preprocessing}
\label{app:data}

All methods read the same processed tensors. Citation uses the supplied dense
document attributes; Twitch expands the supplied feature indices into
3,170-dimensional binary vectors; Airport follows the shared 241-dimensional
one-hot degree-feature construction; and Blog reads the provided
\texttt{attrb} matrices. Features retain the numerical scale supplied by the
processed data.

Edges retain the directed representation received by the trainer, including
stored duplicate edges and self-loops in Airport. Each graph convolution
applies standard GCN normalization and inserts missing self-loops. The local
expert operates on node attributes independently of \texttt{edge\_index};
Airport attributes retain their supplied degree-derived
entries.

\begin{table}[H]
    \centering
    \caption{Statistics of the processed graphs used by all transfer tasks.}
    \label{tab:data-stats}
    \small
    \input{arxiv/tables/dataset_stats}
\end{table}

\subsection{Evaluation and statistical analysis}
\label{app:statistics}

Macro-F1 is the primary metric. All tabulated Macro- and Micro-F1 values are
reported as percentages. Micro@Macro evaluates Micro-F1 at the
Macro-best epoch, while independent Micro-best may select a different epoch.
Macro-best and independent Micro-best coincide in 58 of 80 selected task--run
trajectories and differ in 22. Family and Full-16 scores are equal-task
averages. Controlled contrasts pair task and run. The confidence intervals for
aggregate expert and capacity comparisons use 10,000 two-level paired-bootstrap
replicates that first resample tasks and then paired runs within each task.
We take the 2.5th and 97.5th percentiles of the resulting paired-difference distribution.

\section{Efficiency Analysis}

\subsection{Computational complexity}
\label{app:complexity}

Let $h_g$ and $h_\ell$ denote the graph and local hidden widths. Graph
propagation and linear transformations cost
$O(P_d|E_d|D+|E_d|h_g+N_dDh_g+N_dh_gK)$ in domain $d$. The local expert costs
$O(N_dDh_\ell+N_dh_\ell K)$.
Each of the $R$ EAM repetitions forms pairwise kernels over $2m$ embeddings,
giving $O(Rm^2h_g)$ time and $O(m^2)$ auxiliary memory. EAM is a training-time
operation; inference comprises the graph-aware expert, one local-expert forward pass, and
probability interpolation.

\subsection{Final-model efficiency}
\label{app:efficiency}

Table~\ref{tab:efficiency} measures final-model training and inference under
one matched A$\rightarrow$C configuration on an RTX 4090 D GPU, separately
from the one-time hyperparameter-search cost.

\begin{table}[H]
    \centering
    \caption{Final-model efficiency under a common A$\rightarrow$C profiling setup.}
    \label{tab:efficiency}
    \small
    \input{arxiv/tables/efficiency}
\end{table}

Relative to Graph-only, the complete method adds 50.0\% trainable parameters
while increasing measured training time by 2.3\% and inference latency by
9.4\%; peak memory remains unchanged in this profile. At nearly identical
parameter count, it trains 10.7\% faster than the widened parameter-matched
graph control. It also reduces training and inference time by 48.9\% and
45.9\%, respectively, compared with Graph+Graph. The graph-free local expert therefore
provides heterogeneous evidence at substantially lower execution cost than a
second graph expert.

The operation types help explain this difference between parameter and runtime
overhead. The local expert applies two node-wise dense transformations without
neighborhood aggregation or pairwise alignment kernels. Adding its parameters
therefore does not duplicate the graph expert's propagation and EAM workload.
At inference, combining the two probability vectors requires only $O(N_tK)$
arithmetic operations, with no additional graph traversal. This separation is
consistent with the modest measured latency increase despite the larger
parameter count.

\Needspace{8\baselineskip}
\section{Additional Experiment Results}
\label{app:results}

\paragraph{Consistency across runs.}
Across leave-one-run evaluations, the resulting operating points retain
99.48\% of the Full-16 Macro-F1 score obtained using all five runs. Family-level
scores remain similarly consistent, indicating stable responses across repeated runs.

\subsection{Complete Micro-F1 comparison}
\label{app:micro-results}

Table~\ref{tab:main-results-micro} complements the main Macro-F1 comparison
with Micro-F1 on all 16 transfers. The same task ordering and method grouping
are retained so that class-balanced and frequency-weighted performance can be
examined together under a common transfer setting.

For single-label node classification, Micro-F1 pools true positives, false
positives, and false negatives over classes and equals the fraction of
correctly classified nodes. Macro-F1 instead averages class-specific F1 scores
with equal class weights. The two metrics therefore emphasize different
aspects of prediction when class frequencies are unequal: Micro-F1 summarizes
node-level correctness, while Macro-F1 gives each class equal influence on
the reported score.

\paragraph{Cross-metric consistency.}
\method{} obtains the highest mean Micro-F1 on 15 of 16 transfers, including
all Airport, Blog, and Twitch directions. On Citation, it reaches 83.22 on
A$\rightarrow$C and 79.28 on C$\rightarrow$D; DGSDA retains the strongest
D$\rightarrow$C result at 82.55. The broad agreement with the main Macro-F1
comparison shows that the gains extend to frequency-weighted node prediction,
rather than appearing only under equal weighting of classes.

\paragraph{Family-level behavior.}
The gains over the strongest competing Micro-F1 results are 5.73 and 3.12
points on B1$\rightarrow$B2 and B2$\rightarrow$B1, respectively. Airport
also shows clear improvements, including 4.81 points on U$\rightarrow$B
and 3.90 on B$\rightarrow$E. Twitch gains are smaller: \method{} reaches
60.80 on DE$\rightarrow$EN and 65.82 on EN$\rightarrow$DE. These differences
agree with the main paper's family-dependent gains: additional graph-free
predictions are more useful on some transfers than on others.

\input{arxiv/tables/main_comparison_micro}

\begin{table}[H]
    \centering
    \caption{Complete Micro-F1 comparison (\%) across the 16 transfer tasks.
    Entries report mean $\pm$ sample standard deviation; bold and underline
    denote the best and second-best results.}
    \label{tab:main-results-micro}
    \begingroup
    \setlength{\tabcolsep}{2.2pt}
    \renewcommand{\arraystretch}{1.04}
    \resizebox{\textwidth}{!}{\microtablefirst}
    \par\smallskip
    \resizebox{\textwidth}{!}{\microtablesecond}
    \endgroup
\end{table}

\subsection{Probability quality}
\label{app:controls}

The correctness decomposition in Figure~\ref{fig:complementarity}(a) is
complemented by probability-quality statistics. Mean multiclass Brier scores
are 0.5331 for Graph-only,
0.6512 for Local-only, and 0.5037 after fusion. Mean squared expert
disagreement is 0.2695, the probability-mixture diversity credit is 0.0148,
and the decomposition residual is below $3\times10^{-9}$ in magnitude.
Fusion therefore reduces Brier risk by 0.0293 relative to Graph-only even
though Local-only has higher standalone risk. This pattern is characteristic
of complementary experts: the local probabilities contribute useful class mass
on disagreement nodes, while the graph-dominant mixture preserves the stronger
graph-aware expert on the remaining nodes. The near-zero residual also
confirms the numerical agreement between the observed mixture risk and
Eq.~\ref{eq:brier-decomposition} for the trained expert pair.

\subsection{Matched EAM control}
\label{app:eam-control}

The matched Uniform variant uses the same exact sampler, sample count, kernel,
bandwidth rule, and gradient routes as EAM; uniform target mass
$q_i=1/N_t$ is the controlled intervention. Across the 16 transfer tasks, EAM
improves the fused Macro-F1 endpoint by $0.24$ points on average. The
average gains by family are $+0.35$ (Citation), $+0.06$ (Airport),
$+0.59$ (Blog), and $+0.08$ (Twitch), all measured in Macro-F1 points
relative to the matched control.

\begin{table}[H]
    \centering
    \caption{Per-task EAM gains over matched Uniform sampling (Macro-F1 percentage points).}
    \label{tab:eam-full16}
    \small
    \input{arxiv/tables/eam_full16}
\end{table}

The response follows the quality and coverage of confident target evidence.
Representative improvements include A$\rightarrow$D
($+1.16$ points), B2$\rightarrow$B1 ($+0.80$), C$\rightarrow$D ($+0.43$),
and D$\rightarrow$C ($+0.37$). In these directions, low-entropy predictions
repeatedly provide concentrated anchors for the empirical alignment measure.
More moderate gains arise when the uniform and entropy-tilted samplers already
cover similar target regions. Because the kernel, sample count, and gradient
route are matched, the observed differences directly reflect how
entropy-guided participation changes the target evidence presented to the same
kernel discrepancy estimator.

\subsection{Controlled topology-shift stress test}
\label{app:topology-shift}

The main paper reports graph-expert degradation and local-exclusive evidence. Here we
verify that the intervention reduces target homophily and show the corresponding
fusion response. A$\rightarrow$C, U$\rightarrow$B, B1$\rightarrow$B2, and
EN$\rightarrow$DE each use six degree-preserving rewiring levels and five runs
(120 trajectories); the fusion coefficient remains fixed across severities.

\begin{figure}[H]
    \centering
    \includegraphics[width=0.98\textwidth]{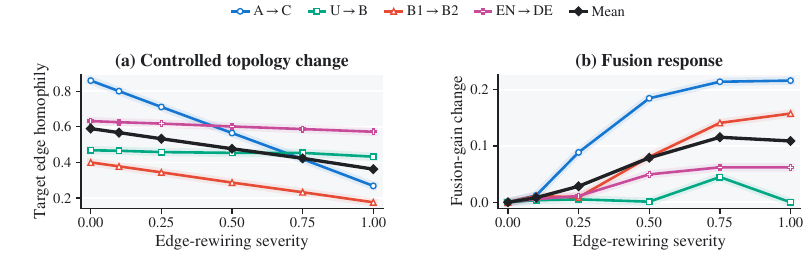}
    \caption{\textbf{Supplementary topology intervention.} Degree-preserving
    rewiring lowers target homophily, while the fusion response varies by graph
    family. The black line is the equal-task mean.}
    \label{fig:topology-shift-supplement}
\end{figure}

Severity zero reproduces the reference endpoint, and local-expert logits have
identical hashes across severities. Table~\ref{tab:topology-endpoints}
summarizes the maximum-severity changes. The largest response occurs on
A$\rightarrow$C: graph damage reaches 69.86 points, local-exclusive correctness
rises by 33.38 points, and the fusion gain increases by 21.62 points.
B1$\rightarrow$B2 and EN$\rightarrow$DE form intermediate regimes, whereas
U$\rightarrow$B remains nearly stable. These trajectories expose a graded
relationship between neighborhood degradation and the value of graph-free local
evidence rather than a binary task split.

\begin{table}[H]
    \centering
    \caption{Endpoint changes at maximum target-topology corruption (percentage points).}
    \label{tab:topology-endpoints}
    \small
    \input{arxiv/tables/topology_endpoints}
\end{table}

Table~\ref{tab:topology-endpoints} reports the primary maximum-severity
endpoints. A complementary fixed-reference-epoch replay yields mean
graph-damage, local-exclusive, and fusion-gain changes of 31.29, 11.67, and
1.91 points, respectively. Both readouts preserve the same ordering: transfers
with stronger graph-expert degradation expose more local-exclusive evidence and
larger gains from combining the two experts.

\subsection{Hyperparameter response}
\label{app:sensitivity}

\begin{figure}[H]
    \centering
    \includegraphics[width=0.96\textwidth]{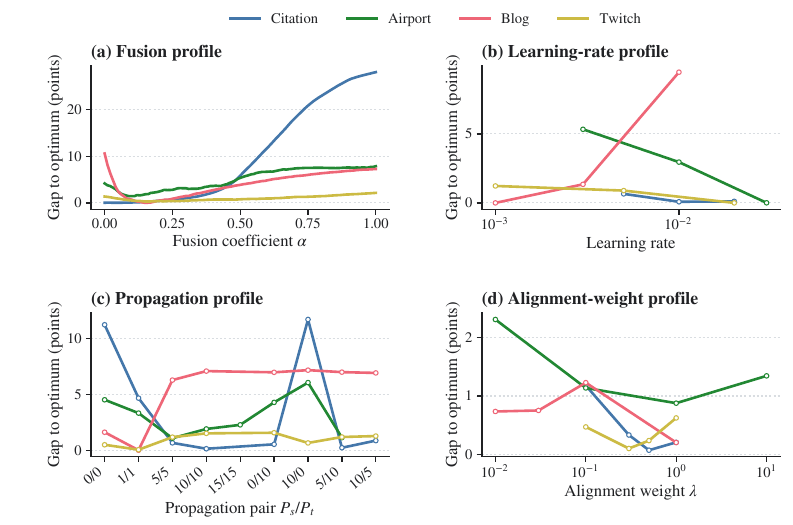}
    \caption{\textbf{Family-level hyperparameter profiles.} Each point reports
    the mean gap to the task optimum after profiling over the remaining search
    dimensions; lower is better.}
    \label{fig:hyperparameter-profiles}
\end{figure}

The search evaluates 96 trained bundles per task and replays the task-level
coefficient on the shared coarse grid
$\alpha_\tau\in\{0.05,0.1,0.4,0.6\}$. Coefficient replay over stored expert
predictions is parameter-free.
Figure~\ref{fig:hyperparameter-profiles} profiles each search dimension while
optimizing over the remaining dimensions; lower gaps indicate wider
high-performing regions across the transfer tasks.

The following analysis relates each parameter response to its role in graph transfer.

\paragraph{Study on the fusion coefficient.}
The coefficient controls the balance between graph-aware and graph-free local
evidence. The shared grid covers graph-dominant fusion, light local-expert
participation, and progressively stronger local-expert contributions. Citation
generally favors the graph-dominant end, whereas Airport, Blog, and Twitch
contain transfers that benefit from larger local weights.  The preferred point
may differ between transfer scenarios, but within a scenario it is one scalar
shared by all nodes. This controlled response supports a compact task-level
probability mixture.

\paragraph{Study on propagation depth.}
Propagation determines how strongly each domain incorporates neighborhood
information before alignment. Symmetric and asymmetric pairs form several
competitive operating regions, with their ordering varying across families.
This agrees with the premise that source and target graphs can require different
degrees of structural smoothing and motivates evaluating source and target
propagation together as a paired configuration.

\paragraph{Study on learning rate and alignment weight.}
Learning-rate profiles retain clear favorable regions for each family. The
alignment weight exhibits the expected balance: increasing it strengthens
cross-domain matching until source discrimination and alignment reach an
effective trade-off. Across both dimensions, the response curves identify
family-level differences in the balance between source fitting and target alignment.

%% file: arxiv/tables/frozen_settings.tex
\begin{tabular*}{0.94\textwidth}{@{\extracolsep{\fill}}lp{0.70\textwidth}}
\toprule
Component & Globally frozen setting \\
\midrule
Experts
& Graph hidden width 128; local hidden width 64; one graph feature layer;
  ReLU; no dropout \\
Optimization
& Adam; local learning-rate scale 1; 150 epochs; no warmup \\
EAM
& Floor $0.8$; source gradient scale $\gamma=0$; sample cap 1000;
  five replacement draws \\
Kernel estimator
& Biased MMD; five RBF kernels; multiplier 2; no cosine mixture \\
Fusion
& Task-level constant mixture in probability space \\
\bottomrule
\end{tabular*}

%% file: arxiv/tables/formal_search_space.tex
\begin{tabular*}{\textwidth}{@{\extracolsep{\fill}}lllll}
\toprule
Family & Graph LR & Propagation pairs $(P_s/P_t)$ & $\lambda$ & WD \\
\midrule
Citation
& $\{.005,.01,.02\}$
& \shortstack[l]{$0/0,1/1,5/5,10/10$\\$0/10,10/0,5/10,10/5$}
& $\{.1,.3,.5,1\}$ & $.001$ \\
Airport
& $\{.003,.01,.03\}$
& \shortstack[l]{$0/0,1/1,5/5,10/10$\\$15/15,0/10,10/0,5/10$}
& $\{.01,.1,1,10\}$ & $.005$ \\
Blog
& $\{.001,.003,.01\}$
& \shortstack[l]{$0/0,1/1,5/5,10/10$\\$0/10,10/0,5/10,10/5$}
& $\{.01,.03,.1,1\}$ & $.005$ \\
Twitch
& $\{.001,.005,.02\}$
& \shortstack[l]{$0/0,1/1,5/5,10/10$\\$0/10,10/0,5/10,10/5$}
& $\{.1,.3,.5,1\}$ & $.005$ \\
\bottomrule
\end{tabular*}

%% file: arxiv/tables/dataset_stats.tex
\begin{tabular*}{0.88\textwidth}{@{\extracolsep{\fill}}llrrrr}
\toprule
Domain & Family & Nodes & Edges & Features & Classes \\
\midrule
BRAZIL & Airport & 131 & 2,148 & 241 & 4 \\
EUROPE & Airport & 399 & 11,990 & 241 & 4 \\
USA & Airport & 1,190 & 27,198 & 241 & 4 \\
\addlinespace[2pt]
Blog1 & Blog & 2,300 & 66,942 & 8,189 & 6 \\
Blog2 & Blog & 2,896 & 107,672 & 8,189 & 6 \\
\addlinespace[2pt]
ACMv9 & Citation & 9,360 & 31,112 & 6,775 & 5 \\
Citationv1 & Citation & 8,935 & 30,196 & 6,775 & 5 \\
DBLPv7 & Citation & 5,484 & 16,234 & 6,775 & 5 \\
\addlinespace[2pt]
DE & Twitch & 9,498 & 153,138 & 3,170 & 2 \\
EN & Twitch & 7,126 & 35,324 & 3,170 & 2 \\
\bottomrule
\end{tabular*}

%% file: arxiv/tables/efficiency.tex
\begin{tabular}{lrrrr}
\toprule
Variant & Parameters & s/epoch & Peak MiB & Inference ms \\
\midrule
Graph-only & 867,973 & 0.710 & 1660 & 3.753 \\
Parameter-matched graph & 1,301,957 & 0.813 & 1854 & 3.756 \\
Graph+Graph & 1,735,946 & 1.421 & 1660 & 7.586 \\
Graph+Local (\method) & 1,301,962 & 0.726 & 1660 & 4.105 \\
\bottomrule
\end{tabular}

%% file: arxiv/tables/main_comparison_micro.tex
\newcommand{\microtablefirst}{%
\begin{tabular}{lcccccccc}
\toprule
Method & A$\to$C & A$\to$D & C$\to$A & C$\to$D & D$\to$A & D$\to$C & B1$\to$B2 & B2$\to$B1 \\
\midrule
\multicolumn{9}{l}{\emph{Source-only}} \\
GCN \venue{ICLR'17}
& \mstd{70.64}{1.51} & \mstd{66.18}{0.69} & \mstd{66.59}{0.94} & \mstd{71.04}{0.80} & \mstd{59.10}{0.71} & \mstd{68.59}{2.37} & \mstd{33.31}{1.97} & \mstd{31.86}{1.50} \\
\midrule
\multicolumn{9}{l}{\emph{Graph domain adaptation}} \\
UDAGCN \venue{WWW'20}
& \mstd{80.11}{1.08} & \mstd{74.43}{1.05} & \mstd{72.90}{1.48} & \mstd{77.88}{1.29} & \mstd{65.29}{2.17} & \mstd{76.18}{1.17} & \mstd{34.99}{2.94} & \mstd{33.44}{1.52} \\
GRADE \venue{AAAI'23}
& \mstd{72.53}{1.66} & \mstd{67.73}{2.26} & \mstd{67.76}{0.36} & \mstd{72.86}{1.00} & \mstd{63.07}{0.36} & \mstd{69.45}{1.06} & \mstd{46.66}{2.32} & \mstd{46.06}{1.26} \\
PairAlign \venue{ICML'24}
& \mstd{61.37}{2.09} & \mstd{60.08}{2.51} & \mstd{55.42}{1.67} & \mstd{61.41}{2.46} & \mstd{53.00}{1.00} & \mstd{58.16}{2.41} & \mstd{41.66}{1.63} & \mstd{44.00}{0.90} \\
GraphAlign \venue{KDD'24}
& \mstd{69.31}{1.13} & \mstd{65.88}{1.46} & \mstd{64.18}{0.82} & \mstd{69.38}{1.24} & \mstd{59.46}{0.88} & \mstd{65.17}{0.98} & \mstd{39.73}{1.92} & \mstd{43.78}{1.04} \\
A2GNN-MMD \venue{AAAI'24}
& \mstd{80.82}{0.35} & \mstd{76.05}{0.26} & \mstd{75.68}{0.32} & \mstd{76.39}{1.12} & \mstd{\underline{73.59}}{0.45} & \mstd{80.14}{0.59} & \mstd{40.99}{1.44} & \mstd{42.94}{1.59} \\
TDSS-RW \venue{AAAI'25}
& \mstd{81.81}{0.40} & \mstd{77.55}{0.51} & \mstd{74.59}{0.45} & \mstd{76.55}{0.41} & \mstd{73.15}{0.58} & \mstd{80.49}{0.05} & \mstd{45.00}{0.92} & \mstd{43.99}{0.90} \\
GAA \venue{ICLR'25}
& \mstd{75.56}{1.57} & \mstd{68.17}{1.02} & \mstd{74.77}{0.78} & \mstd{68.25}{1.04} & \mstd{68.97}{0.91} & \mstd{71.21}{0.59} & \mstd{46.11}{1.43} & \mstd{47.77}{1.70} \\
DGSDA \venue{ICML'25}
& \mstd{\underline{82.88}}{0.30} & \mstd{76.15}{0.68} & \mstd{74.70}{0.22} & \mstd{77.67}{0.35} & \mstd{73.20}{0.36} & \mstd{\textbf{82.55}}{0.49} & \mstd{38.17}{1.46} & \mstd{41.98}{1.38} \\
HGDA \venue{ICML'25}
& \mstd{71.36}{1.09} & \mstd{72.54}{1.77} & \mstd{66.62}{1.91} & \mstd{73.05}{1.57} & \mstd{67.57}{1.74} & \mstd{74.99}{1.20} & \mstd{42.47}{0.79} & \mstd{43.96}{1.04} \\
ADAlign \venue{ICLR'26}
& \mstd{81.54}{0.50} & \mstd{\underline{77.59}}{0.54} & \mstd{\underline{75.96}}{0.37} & \mstd{\underline{78.22}}{1.23} & \mstd{70.70}{1.14} & \mstd{78.43}{1.18} & \mstd{44.85}{1.59} & \mstd{44.59}{1.40} \\
DiffGDA \venue{ICLR'26}
& \mstd{79.55}{1.20} & \mstd{74.11}{1.36} & \mstd{73.15}{0.94} & \mstd{77.41}{0.79} & \mstd{68.67}{1.35} & \mstd{77.22}{0.46} & \mstd{42.10}{1.32} & \mstd{41.10}{1.74} \\
DFT \venue{KDD'26}
& \mstd{74.45}{1.81} & \mstd{72.84}{0.49} & \mstd{68.35}{0.79} & \mstd{73.09}{1.00} & \mstd{66.53}{1.34} & \mstd{78.20}{1.59} & \mstd{\underline{48.31}}{1.42} & \mstd{\underline{51.34}}{0.93} \\
\midrule
\rowcolor{oursrow}
\textbf{\method{} (Ours)}
& \mstd{\textbf{83.22}}{0.14} & \mstd{\textbf{78.50}}{0.77} & \mstd{\textbf{76.06}}{0.19} & \mstd{\textbf{79.28}}{0.23} & \mstd{\textbf{73.94}}{0.27} & \mstd{\underline{82.18}}{0.40} & \mstd{\textbf{54.04}}{1.73} & \mstd{\textbf{54.46}}{0.63} \\
\bottomrule
\end{tabular}%
}

\newcommand{\microtablesecond}{%
\begin{tabular}{lcccccccc}
\toprule
Method & U$\to$B & U$\to$E & B$\to$U & B$\to$E & E$\to$U & E$\to$B & DE$\to$EN & EN$\to$DE \\
\midrule
\multicolumn{9}{l}{\emph{Source-only}} \\
GCN \venue{ICLR'17}
& \mstd{50.38}{1.05} & \mstd{37.69}{1.27} & \mstd{44.08}{0.47} & \mstd{36.39}{1.43} & \mstd{45.36}{0.26} & \mstd{42.14}{0.98} & \mstd{56.85}{0.42} & \mstd{59.78}{1.09} \\
\midrule
\multicolumn{9}{l}{\emph{Graph domain adaptation}} \\
UDAGCN \venue{WWW'20}
& \mstd{62.25}{0.85} & \mstd{44.35}{0.93} & \mstd{41.82}{0.66} & \mstd{51.62}{0.93} & \mstd{42.18}{0.64} & \mstd{61.37}{1.16} & \mstd{58.45}{0.64} & \mstd{63.15}{0.53} \\
GRADE \venue{AAAI'23}
& \mstd{62.44}{1.51} & \mstd{48.87}{1.81} & \mstd{42.20}{1.64} & \mstd{55.44}{1.14} & \mstd{46.47}{1.77} & \mstd{70.08}{1.74} & \mstd{59.25}{0.34} & \mstd{63.99}{0.31} \\
PairAlign \venue{ICML'24}
& \mstd{70.00}{1.76} & \mstd{41.45}{0.65} & \mstd{47.41}{1.26} & \mstd{40.45}{1.21} & \mstd{43.38}{1.17} & \mstd{49.62}{1.29} & \mstd{60.28}{0.40} & \mstd{\underline{65.23}}{0.33} \\
GraphAlign \venue{KDD'24}
& \mstd{69.08}{0.87} & \mstd{\underline{57.44}}{0.80} & \mstd{49.82}{0.51} & \mstd{\underline{55.75}}{0.85} & \mstd{\underline{52.12}}{0.47} & \mstd{69.66}{0.60} & \mstd{54.67}{1.20} & \mstd{59.78}{1.86} \\
A2GNN-MMD \venue{AAAI'24}
& \mstd{61.98}{0.72} & \mstd{51.63}{1.42} & \mstd{44.62}{1.48} & \mstd{54.04}{0.29} & \mstd{44.13}{0.86} & \mstd{66.26}{1.35} & \mstd{57.24}{0.37} & \mstd{59.74}{1.25} \\
TDSS-RW \venue{AAAI'25}
& \mstd{75.31}{2.25} & \mstd{45.25}{0.34} & \mstd{56.10}{6.49} & \mstd{44.35}{2.99} & \mstd{51.24}{3.58} & \mstd{62.93}{0.05} & \mstd{56.45}{0.50} & \mstd{57.71}{0.69} \\
GAA \venue{ICLR'25}
& \mstd{75.42}{1.39} & \mstd{54.74}{1.62} & \mstd{55.62}{1.49} & \mstd{54.84}{1.08} & \mstd{50.82}{1.23} & \mstd{\underline{74.20}}{1.77} & \mstd{55.38}{1.36} & \mstd{60.85}{0.99} \\
DGSDA \venue{ICML'25}
& \mstd{63.97}{1.47} & \mstd{53.88}{1.19} & \mstd{57.71}{0.22} & \mstd{53.13}{0.77} & \mstd{52.06}{0.62} & \mstd{71.45}{1.16} & \mstd{60.35}{0.14} & \mstd{63.43}{0.65} \\
HGDA \venue{ICML'25}
& \mstd{75.40}{0.78} & \mstd{44.00}{0.50} & \mstd{\underline{57.80}}{0.64} & \mstd{53.18}{0.28} & \mstd{47.61}{0.20} & \mstd{66.05}{0.39} & \mstd{56.62}{1.02} & \mstd{60.67}{1.06} \\
ADAlign \venue{ICLR'26}
& \mstd{75.57}{1.23} & \mstd{53.73}{1.06} & \mstd{49.73}{1.27} & \mstd{55.69}{1.13} & \mstd{51.04}{1.03} & \mstd{69.16}{1.20} & \mstd{59.76}{0.28} & \mstd{64.58}{0.44} \\
DiffGDA \venue{ICLR'26}
& \mstd{76.26}{1.06} & \mstd{48.07}{1.47} & \mstd{45.70}{1.25} & \mstd{45.92}{0.85} & \mstd{47.76}{1.20} & \mstd{65.34}{1.05} & \mstd{56.37}{0.30} & \mstd{59.66}{0.83} \\
DFT \venue{KDD'26}
& \mstd{\underline{76.56}}{1.02} & \mstd{51.33}{0.75} & \mstd{53.24}{1.21} & \mstd{52.78}{1.55} & \mstd{50.03}{0.89} & \mstd{69.31}{1.55} & \mstd{\underline{60.73}}{0.60} & \mstd{64.73}{1.16} \\
\midrule
\rowcolor{oursrow}
\textbf{\method{} (Ours)}
& \mstd{\textbf{81.37}}{0.42} & \mstd{\textbf{58.25}}{0.76} & \mstd{\textbf{58.32}}{0.39} & \mstd{\textbf{59.65}}{0.18} & \mstd{\textbf{52.69}}{0.41} & \mstd{\textbf{76.64}}{1.16} & \mstd{\textbf{60.80}}{0.13} & \mstd{\textbf{65.82}}{0.57} \\
\bottomrule
\end{tabular}%
}

%% file: arxiv/tables/eam_full16.tex
\begin{tabular}{lr@{\hspace{26pt}}lr}
\toprule
Task & $\Delta$ Macro-F1 & Task & $\Delta$ Macro-F1 \\
\midrule
A$\rightarrow$C  & $+0.03$ & B$\rightarrow$U  & $+0.02$ \\
A$\rightarrow$D  & $+1.16$ & E$\rightarrow$U  & $-0.03$ \\
C$\rightarrow$A  & $+0.17$ & U$\rightarrow$B  & $+0.17$ \\
C$\rightarrow$D  & $+0.43$ & U$\rightarrow$E  & $+0.07$ \\
D$\rightarrow$A  & $-0.04$ & B1$\rightarrow$B2 & $+0.39$ \\
D$\rightarrow$C  & $+0.37$ & B2$\rightarrow$B1 & $+0.80$ \\
B$\rightarrow$E  & $-0.03$ & DE$\rightarrow$EN & $+0.20$ \\
E$\rightarrow$B  & $+0.17$ & EN$\rightarrow$DE & $-0.04$ \\
\bottomrule
\end{tabular}

%% file: arxiv/tables/topology_endpoints.tex
\begin{tabular}{lrrr}
\toprule
Transfer & Graph damage & Local-exclusive gain & Fusion-gain change \\
\midrule
A$\rightarrow$C & 69.86 & 33.38 & 21.62 \\
U$\rightarrow$B & 0.76 & 0.76 & $-0.00$ \\
B1$\rightarrow$B2 & 27.10 & 14.27 & 15.77 \\
EN$\rightarrow$DE & 10.36 & 8.12 & 6.19 \\
\midrule
Equal-task mean & 27.02 & 14.13 & 10.90 \\
\bottomrule
\end{tabular}

%% file: main.bbl
\begin{thebibliography}{36}
\providecommand{\natexlab}[1]{#1}
\providecommand{\url}[1]{\texttt{#1}}
\expandafter\ifx\csname urlstyle\endcsname\relax
  \providecommand{\doi}[1]{doi: #1}\else
  \providecommand{\doi}{doi: \begingroup \urlstyle{rm}\Url}\fi

\bibitem[Ben-David et~al.(2010)Ben-David, Blitzer, Crammer, Kulesza, Pereira,
  and Vaughan]{bendavid2010theory}
Shai Ben-David, John Blitzer, Koby Crammer, Alex Kulesza, Fernando Pereira, and
  Jennifer~Wortman Vaughan.
\newblock A theory of learning from different domains.
\newblock \emph{Machine Learning}, 79\penalty0 (1--2):\penalty0 151--175, 2010.

\bibitem[Cai et~al.(2024)Cai, Wu, Li, Wei, Yi, and Zhang]{cai2024dgda}
Ruichu Cai, Fengzhu Wu, Zijian Li, Pengfei Wei, Lingling Yi, and Kun Zhang.
\newblock Graph domain adaptation: A generative view.
\newblock \emph{ACM Transactions on Knowledge Discovery from Data}, 18\penalty0
  (3), 2024.

\bibitem[Chen et~al.(2025)Chen, Ye, Wang, Zhang, Zhang, Wang, Zhang, and
  Zhuang]{chen2025tdss}
Wei Chen, Guo Ye, Yakun Wang, Zhao Zhang, Libang Zhang, Daixin Wang, Zhiqiang
  Zhang, and Fuzhen Zhuang.
\newblock Smoothness really matters: A simple yet effective approach for
  unsupervised graph domain adaptation.
\newblock In \emph{Proceedings of the AAAI Conference on Artificial
  Intelligence}, volume~39, pp.\  15875--15883, 2025.

\bibitem[Chen et~al.(2026{\natexlab{a}})Chen, Guo, Li, Zhang, Zhong, Zhuang,
  and Wang]{chen2026adalign}
Wei Chen, Xingyu Guo, Shuang Li, Zhao Zhang, Yan Zhong, Fuzhen Zhuang, and
  Deqing Wang.
\newblock Learning adaptive distribution alignment with neural characteristic
  function for graph domain adaptation.
\newblock In \emph{International Conference on Learning Representations},
  2026{\natexlab{a}}.

\bibitem[Chen et~al.(2026{\natexlab{b}})Chen, Guo, Li, Zhong, Zhang, Zhuang,
  Liu, Zhang, Ye, and He]{chen2026diffgda}
Wei Chen, Xingyu Guo, Shuang Li, Yan Zhong, Zhao Zhang, Fuzhen Zhuang, Hongrui
  Liu, Libang Zhang, Guo Ye, and Huimei He.
\newblock Learning structure-semantic evolution trajectories for graph domain
  adaptation.
\newblock In \emph{International Conference on Learning Representations},
  2026{\natexlab{b}}.

\bibitem[Dai et~al.(2023)Dai, Wu, Xiao, Shen, and Wang]{dai2023adagcn}
Quanyu Dai, Xiao-Ming Wu, Jiaren Xiao, Xiao Shen, and Dan Wang.
\newblock Graph transfer learning via adversarial domain adaptation with graph
  convolution.
\newblock \emph{IEEE Transactions on Knowledge and Data Engineering},
  35\penalty0 (5):\penalty0 4908--4922, 2023.

\bibitem[Fang et~al.(2025{\natexlab{a}})Fang, Li, Kang, Zeng, Dashtbayaz, Pu,
  Wang, and Ling]{fang2025attrgda}
Ruiyi Fang, Bingheng Li, Zhao Kang, Qiuhao Zeng, Nima~Hosseini Dashtbayaz,
  Ruizhi Pu, Boyu Wang, and Charles Ling.
\newblock On the benefits of attribute-driven graph domain adaptation.
\newblock In \emph{International Conference on Learning Representations},
  2025{\natexlab{a}}.

\bibitem[Fang et~al.(2025{\natexlab{b}})Fang, Li, Zhao, Pu, Zeng, Xu, Ling, and
  Wang]{fang2025hgda}
Ruiyi Fang, Bingheng Li, Jingyu Zhao, Ruizhi Pu, Qiuhao Zeng, Gezheng Xu,
  Charles Ling, and Boyu Wang.
\newblock Homophily enhanced graph domain adaptation.
\newblock In \emph{Proceedings of the 42nd International Conference on Machine
  Learning}, volume 267 of \emph{Proceedings of Machine Learning Research},
  pp.\  16006--16028, 2025{\natexlab{b}}.

\bibitem[Ganin et~al.(2016)Ganin, Ustinova, Ajakan, Germain, Larochelle,
  Laviolette, Marchand, and Lempitsky]{ganin2016dann}
Yaroslav Ganin, Evgeniya Ustinova, Hana Ajakan, Pascal Germain, Hugo
  Larochelle, Fran\c{c}ois Laviolette, Mario Marchand, and Victor Lempitsky.
\newblock Domain-adversarial training of neural networks.
\newblock \emph{Journal of Machine Learning Research}, 17\penalty0
  (59):\penalty0 1--35, 2016.

\bibitem[Gneiting \& Raftery(2007)Gneiting and Raftery]{gneiting2007proper}
Tilmann Gneiting and Adrian~E. Raftery.
\newblock Strictly proper scoring rules, prediction, and estimation.
\newblock \emph{Journal of the American Statistical Association}, 102\penalty0
  (477):\penalty0 359--378, 2007.

\bibitem[Goldfeld \& Polyanskiy(2020)Goldfeld and
  Polyanskiy]{goldfeld2020information}
Ziv Goldfeld and Yury Polyanskiy.
\newblock The information bottleneck problem and its applications in machine
  learning.
\newblock \emph{IEEE Journal on Selected Areas in Information Theory},
  1\penalty0 (1):\penalty0 19--38, 2020.

\bibitem[Gretton et~al.(2012)Gretton, Borgwardt, Rasch, Sch\"olkopf, and
  Smola]{gretton2012kernel}
Arthur Gretton, Karsten~M. Borgwardt, Malte~J. Rasch, Bernhard Sch\"olkopf, and
  Alexander Smola.
\newblock A kernel two-sample test.
\newblock \emph{Journal of Machine Learning Research}, 13:\penalty0 723--773,
  2012.

\bibitem[Huang et~al.(2024)Huang, Xu, Jiang, An, and Yang]{huang2024graphalign}
Renhong Huang, Jiarong Xu, Xin Jiang, Ruichuan An, and Yang Yang.
\newblock Can modifying data address graph domain adaptation?
\newblock In \emph{Proceedings of the 30th ACM SIGKDD Conference on Knowledge
  Discovery and Data Mining}, pp.\  1131--1142, 2024.

\bibitem[Jacobs et~al.(1991)Jacobs, Jordan, Nowlan, and Hinton]{jacobs1991moe}
Robert~A. Jacobs, Michael~I. Jordan, Steven~J. Nowlan, and Geoffrey~E. Hinton.
\newblock Adaptive mixtures of local experts.
\newblock \emph{Neural Computation}, 3\penalty0 (1):\penalty0 79--87, 1991.

\bibitem[Ju et~al.(2025)Ju, Yang, Li, and Wang]{ju2025graphbridge}
Li~Ju, Xingyi Yang, Qi~Li, and Xinchao Wang.
\newblock Graphbridge: Towards arbitrary transfer learning in gnns.
\newblock In \emph{International Conference on Learning Representations}, 2025.

\bibitem[Kipf \& Welling(2017)Kipf and Welling]{kipf2017gcn}
Thomas~N. Kipf and Max Welling.
\newblock Semi-supervised classification with graph convolutional networks.
\newblock In \emph{International Conference on Learning Representations}, 2017.

\bibitem[Krogh \& Vedelsby(1994)Krogh and Vedelsby]{krogh1994ensembles}
Anders Krogh and Jesper Vedelsby.
\newblock Neural network ensembles, cross validation, and active learning.
\newblock In \emph{Advances in Neural Information Processing Systems},
  volume~7, 1994.

\bibitem[Lakshminarayanan et~al.(2017)Lakshminarayanan, Pritzel, and
  Blundell]{lakshminarayanan2017deepensembles}
Balaji Lakshminarayanan, Alexander Pritzel, and Charles Blundell.
\newblock Simple and scalable predictive uncertainty estimation using deep
  ensembles.
\newblock In \emph{Advances in Neural Information Processing Systems},
  volume~30, 2017.

\bibitem[Liu et~al.(2024{\natexlab{a}})Liu, Fang, Zhang, Gu, Zhou, Wang, and
  Bu]{liu2024a2gnn}
Meihan Liu, Zeyu Fang, Zhen Zhang, Ming Gu, Sheng Zhou, Xin Wang, and Jiajun
  Bu.
\newblock Rethinking propagation for unsupervised graph domain adaptation.
\newblock In \emph{Proceedings of the AAAI Conference on Artificial
  Intelligence}, volume~38, pp.\  13963--13971, 2024{\natexlab{a}}.

\bibitem[Liu et~al.(2023)Liu, Li, Feng, Tran, Zhao, Qiu, and Li]{liu2023struRW}
Shikun Liu, Tianchun Li, Yongbin Feng, Nhan Tran, Han Zhao, Qiang Qiu, and Pan
  Li.
\newblock Structural re-weighting improves graph domain adaptation.
\newblock In \emph{Proceedings of the 40th International Conference on Machine
  Learning}, volume 202 of \emph{Proceedings of Machine Learning Research},
  pp.\  21778--21793, 2023.

\bibitem[Liu et~al.(2024{\natexlab{b}})Liu, Zou, Zhao, and
  Li]{liu2024pairalign}
Shikun Liu, Deyu Zou, Han Zhao, and Pan Li.
\newblock Pairwise alignment improves graph domain adaptation.
\newblock In \emph{Proceedings of the 41st International Conference on Machine
  Learning}, volume 235 of \emph{Proceedings of Machine Learning Research},
  pp.\  32552--32575, 2024{\natexlab{b}}.

\bibitem[Long et~al.(2015)Long, Cao, Wang, and Jordan]{long2015dan}
Mingsheng Long, Yue Cao, Jianmin Wang, and Michael~I. Jordan.
\newblock Learning transferable features with deep adaptation networks.
\newblock In \emph{Proceedings of the 32nd International Conference on Machine
  Learning}, volume~37 of \emph{Proceedings of Machine Learning Research}, pp.\
   97--105, 2015.

\bibitem[Pang et~al.(2023)Pang, Wang, Tang, Xiao, and Yin]{pang2023sagda}
Jinhui Pang, Zixuan Wang, Jiliang Tang, Mingyan Xiao, and Nan Yin.
\newblock {SA-GDA}: Spectral augmentation for graph domain adaptation.
\newblock In \emph{Proceedings of the 31st ACM International Conference on
  Multimedia}, pp.\  309--318, 2023.

\bibitem[Shazeer et~al.(2017)Shazeer, Mirhoseini, Maziarz, Davis, Le, Hinton,
  and Dean]{shazeer2017moe}
Noam Shazeer, Azalia Mirhoseini, Krzysztof Maziarz, Andy Davis, Quoc~V. Le,
  Geoffrey~E. Hinton, and Jeff Dean.
\newblock Outrageously large neural networks: The sparsely-gated
  mixture-of-experts layer.
\newblock In \emph{International Conference on Learning Representations}, 2017.

\bibitem[Shen et~al.(2020)Shen, Dai, Chung, Lu, and Choi]{shen2020acdne}
Xiao Shen, Quanyu Dai, Fu-lai Chung, Wei Lu, and Kup-Sze Choi.
\newblock Adversarial deep network embedding for cross-network node
  classification.
\newblock In \emph{Proceedings of the AAAI Conference on Artificial
  Intelligence}, volume~34, pp.\  2991--2999, 2020.

\bibitem[Sun \& Saenko(2016)Sun and Saenko]{sun2016deepcoral}
Baochen Sun and Kate Saenko.
\newblock Deep {CORAL}: Correlation alignment for deep domain adaptation.
\newblock In \emph{European Conference on Computer Vision Workshops}, pp.\
  443--450, 2016.

\bibitem[Tai et~al.(2026)Tai, Zou, and Wang]{tai2026dft}
Xinwei Tai, Dongmian Zou, and Hongfei Wang.
\newblock Enhancing node-level graph domain adaptation by alleviating local
  dependency.
\newblock In \emph{Proceedings of the 32nd ACM SIGKDD Conference on Knowledge
  Discovery and Data Mining}, pp.\  1366--1377, 2026.

\bibitem[Tzeng et~al.(2017)Tzeng, Hoffman, Saenko, and Darrell]{tzeng2017adda}
Eric Tzeng, Judy Hoffman, Kate Saenko, and Trevor Darrell.
\newblock Adversarial discriminative domain adaptation.
\newblock In \emph{Proceedings of the IEEE Conference on Computer Vision and
  Pattern Recognition}, pp.\  7167--7176, 2017.

\bibitem[Wu et~al.(2023)Wu, He, and Ainsworth]{wu2023grade}
Jun Wu, Jingrui He, and Elizabeth~A. Ainsworth.
\newblock Non-iid transfer learning on graphs.
\newblock In \emph{Proceedings of the AAAI Conference on Artificial
  Intelligence}, volume~37, pp.\  10342--10350, 2023.

\bibitem[Wu et~al.(2020)Wu, Pan, Zhou, Chang, and Zhu]{wu2020udagcn}
Man Wu, Shirui Pan, Chuan Zhou, Xiaojun Chang, and Xingquan Zhu.
\newblock Unsupervised domain adaptive graph convolutional networks.
\newblock In \emph{Proceedings of The Web Conference 2020}, pp.\  1457--1467,
  2020.

\bibitem[Yang et~al.(2025)Yang, Chen, Zhuo, Jin, Wang, Cao, Wang, and
  Guo]{yang2025dgsda}
Liang Yang, Xin Chen, Jiaming Zhuo, Di~Jin, Chuan Wang, Xiaochun Cao, Zhen
  Wang, and Yuanfang Guo.
\newblock Disentangled graph spectral domain adaptation.
\newblock In \emph{Proceedings of the 42nd International Conference on Machine
  Learning}, volume 267 of \emph{Proceedings of Machine Learning Research},
  pp.\  70632--70648, 2025.

\bibitem[Yang et~al.(2024)Yang, Wang, Yu, He, Huang, and Jin]{yang2024jdagcn}
Niya Yang, Ye~Wang, Zhizhi Yu, Dongxiao He, Xin Huang, and Di~Jin.
\newblock Joint domain adaptive graph convolutional network.
\newblock In \emph{Proceedings of the Thirty-Third International Joint
  Conference on Artificial Intelligence}, pp.\  2496--2504, 2024.

\bibitem[You et~al.(2023)You, Chen, Wang, and Shen]{you2023specreg}
Yuning You, Tianlong Chen, Zhangyang Wang, and Yang Shen.
\newblock Graph domain adaptation via theory-grounded spectral regularization.
\newblock In \emph{International Conference on Learning Representations}, 2023.

\bibitem[Zeng et~al.(2024)Zeng, Lyu, Hu, Xia, and Luo]{zeng2024mowst}
Hanqing Zeng, Hanjia Lyu, Diyi Hu, Yinglong Xia, and Jiebo Luo.
\newblock Mixture of weak and strong experts on graphs.
\newblock In \emph{International Conference on Learning Representations}, 2024.

\bibitem[Zhang et~al.(2019)Zhang, Song, Du, Yang, and Jin]{zhang2019dane}
Yizhou Zhang, Guojie Song, Lun Du, Shuwen Yang, and Yilun Jin.
\newblock {DANE}: Domain adaptive network embedding.
\newblock In \emph{Proceedings of the Twenty-Eighth International Joint
  Conference on Artificial Intelligence}, pp.\  4362--4368, 2019.

\bibitem[Zhu et~al.(2020)Zhu, Yan, Zhao, Heimann, Akoglu, and
  Koutra]{zhu2020h2gcn}
Jiong Zhu, Yujun Yan, Lingxiao Zhao, Mark Heimann, Leman Akoglu, and Danai
  Koutra.
\newblock Beyond homophily in graph neural networks: Current limitations and
  effective designs.
\newblock In \emph{Advances in Neural Information Processing Systems},
  volume~33, pp.\  7793--7804, 2020.

\end{thebibliography}
